\documentclass[preprint, review,3p,times, 10pt]{elsarticle}

\usepackage{cite}
\usepackage{amsmath,amssymb,amsfonts}
\usepackage{graphicx}
\usepackage{textcomp}
\usepackage{comment}
\usepackage{footnote}
\usepackage{comment}
\usepackage{threeparttable}
\usepackage{romannum}
\usepackage{multirow}
\usepackage{makecell}
\usepackage{longtable}
\usepackage{lscape}
\usepackage{soul}
\usepackage{array, makecell}
\usepackage{pifont}

\usepackage{algorithm}
\usepackage{algpseudocode}

\usepackage{longtable}

\usepackage[hyphens]{url}
\usepackage{hyperref}
\hypersetup{colorlinks=true,breaklinks=true}

\usepackage{natbib}
\usepackage{multirow}
\usepackage{array, makecell}

\usepackage{pdflscape}   
\usepackage{booktabs}    

\usepackage{xcolor}
\usepackage{pifont}

\usepackage{pgfplots}
\pgfplotsset{compat=newest} 
\usetikzlibrary{patterns}

\usepackage{caption}
\usepackage{subcaption}

\journal{Elsevier}

\begin{document}

\begin{frontmatter}
\title{Community-Based Early-Stage Chronic Kidney Disease Screening using Explainable Machine Learning for Low-Resource Settings}

\author[1]{Muhammad Ashad Kabir \corref{correspondingauthor}}
\cortext[correspondingauthor]{Corresponding author: Charles Sturt University, Panorama Ave, Bathurst, NSW 2795. akabir@csu.edu.au}
\ead{akabir@csu.edu.au}%
\author[2]{Sirajam Munira}\ead{munirs@rpi.edu}
\author[3]{Dewan Tasnia Azad}\ead{dewan.azad@icddrb.org}
\author[3]{Saleh Mohammed Ikram}\ead{saleh.ikram@icddrb.org}
\author[4]{Mohammad Habibur Rahman Sarker}\ead{habibur.rahman@icddrb.org}
\author[3]{Syed Manzoor Ahmed Hanifi}\ead{hanifi@icddrb.org}
\affiliation[1]{organization={School of Computing, Mathematics and Engineering, Charles Sturt University}, city={Bathurst}, state={NSW}, postcode={2795}, country={Australia}}
\affiliation[2]{organization={Department of Computer Science, Rensselaer Polytechnic Institute}, city={Troy}, state={NY},postcode={12180},country={USA}}

\affiliation[3]{organization={Health Systems and Population Studies Division, International Centre for Diarrhoeal Disease Research, Bangladesh (icddr,b)}, city={Dhaka}, postcode={1212},country={Bangladesh}}

\affiliation[4]{organization={Technical Training Unit, International Centre for Diarrhoeal Disease Research, Bangladesh (icddr,b)}, city={Dhaka}, postcode={1212},country={Bangladesh}}

\begin{abstract}
\textbf{Background:} Early detection of chronic kidney disease (CKD) is essential for preventing progression to end-stage renal disease. However, existing screening tools—primarily developed using populations from high-income countries—often underperform in Bangladesh and South Asia, where risk profiles differ. Moreover, many machine learning–based CKD studies rely heavily on pathology-test variables and are developed using hospital-based datasets, limiting their applicability for community-level screening and for settings where laboratory testing is not readily accessible.
 
\textbf{Objective:} To develop and evaluate an explainable machine learning (ML) framework for community-based early-stage CKD screening that derives and evaluates optimized \textit{predictor subsets} from both all available variables and variables excluding pathology tests, enabling accurate and practical risk assessment in low-resource settings.

\textbf{Methods:} A community-based CKD dataset from Bangladesh was used to develop predictive models. Variables were organized into clinically meaningful feature groups, and ten complementary feature selection methods were applied to identify robust predictor subsets. Twelve ML classifiers were evaluated using nested cross-validation. Model performance was benchmarked against established CKD screening tools and externally validated on three independent datasets from India, the UAE, and Bangladesh. SHAP (SHapley Additive exPlanations) was used to interpret model predictions.

\textbf{Results:} The best-performing model using the optimized feature subset derived from all variables achieved a balanced accuracy of 90.4\%, while the subset derived from variables excluding pathology tests achieved comparable performance (89.23\%), demonstrating the feasibility of accurate CKD risk prediction using minimal non-laboratory features. The proposed approach outperformed existing screening tools while requiring fewer and more accessible inputs. External validation demonstrated strong generalizability, with sensitivities ranging from 78\% to 98\%. SHAP analysis identified clinically meaningful predictors consistent with established CKD risk factors.

\textbf{Conclusions:} Accurate, interpretable, and scalable early-stage CKD screening is achievable using only non-pathology-test features. This framework demonstrates the potential for community-level CKD screening in resource-constrained settings. 

\end{abstract}

\begin{keyword}
Chronic kidney disease \sep early stage \sep machine learning \sep explainable \sep feature selection \sep community-based \sep screening
\end{keyword}
\end{frontmatter}

\section{Introduction}
\label{sec:introduction}

Chronic kidney disease (CKD) represents a critical and growing global health challenge, with profound socioeconomic implications. Globally, kidney disease affects approximately 850 million people~\citep{jager2019single}, with CKD estimated to have a prevalence of 11–13\%~\citep{hill2016global}. In 2017, around 1.2 million people died from CKD, and it was the 12th leading cause of death worldwide~\citep{bikbov2020global}. The overall number of CKD-related deaths has increased substantially over the recent decade, yet resources for managing this silent epidemic remain limited in most developing regions~\citep{bikbov2020global}. In low- and middle-income countries (LMICs), including South Asia, the burden is particularly severe due to limited healthcare infrastructure, shortage of nephrology care, and poor access to diagnostics~\citep{abraham2016chronic}. In Bangladesh, community-based studies in urban and rural populations suggest a prevalence rate of CKD 26\% and 22\%, respectively~\citep{anand2014high,sarker2021community,banik2021prevalence}, with most patients unaware of their condition~\citep{iqbal2018knowledge}.

Early-stage CKD detection is essential to delay or prevent the onset of end-stage kidney disease (ESKD), reduce cardiovascular complications, and avoid the substantial costs associated with renal replacement therapies (RRT). Evidence suggests that CKD progression can be slowed considerably through early lifestyle modification, pharmacological management of comorbidities, and patient education \citep{pollock2024framework}. However, early-stage CKD is typically asymptomatic; only about 4\% of individuals with CKD stages~1--2 and fewer than 10\% with stage~3 are aware of their condition~\citep{plantinga2008patient}, and most patients are diagnosed only after clinical symptoms emerge, by which time renal impairment is often irreversible \citep{levey2022chronic}. The failure to detect CKD at an early stage not only worsens patient outcomes but also imposes a significant economic burden on health systems~\citep{luyckx2021global}, particularly in low-income settings where RRT is prohibitively expensive and largely inaccessible \citep{niang2018hemodialysis}. 

Standard clinical diagnosis of CKD relies on laboratory or pathology tests such as serum creatinine levels, estimated glomerular filtration rate (eGFR), and the urinary albumin-to-creatinine ratio (ACR) \citep{levin2024executive}. These assessments require access to well-equipped diagnostic facilities, trained personnel, and regular follow-up---resources that are scarce in many low-resource settings \citep{stanifer2014chronic}. Early diagnosis is particularly challenging in the Global South, including Bangladesh, where a large proportion of the population lives in rural and peri-urban communities with limited access to healthcare infrastructure, diagnostic laboratories, and nephrology specialists \citep{kabir2023capacity}. In addition, poor awareness among patients and providers, inadequate communication between them, low risk perception among high-risk individuals, and the financial burden associated with travel, laboratory testing, and clinical consultations further deters early CKD diagnosis~\citep{zeba2020early,jafar2020access,nazar2014barriers,kahn2015chronic}. Together, these structural and socioeconomic barriers contribute to delayed recognition of CKD and missed opportunities for timely intervention.

Several community-based CKD screening tools (e.g., SCORED~\citep{bang2007development}) have been developed over the past two decades to facilitate risk assessment~\citep{stolpe2022external}; however, most were designed in high-income countries and exhibit limited relevance for low-resource settings (Table~\ref{tab:sota} provides a summary). Prominent tools include the SCORED~\citep{bang2007development} from the United States, \citet{kshirsagar2008simple} based on the U.S. population, the \citet{thakkinstian2011simple} developed in Thailand, the \citet{kearns2013predicting} from England, and \citet{kwon2019simple} from South Korea. These tools rely largely on demographic and medical history variables such as age, sex, diabetes, hypertension, and cardiovascular disease, with several incorporating clinical examination features (e.g., anemia, blood pressure, dipstick proteinuria). Importantly, most were designed to detect CKD stages 3--5, thereby overlooking early-stage disease where intervention is most impactful. Their dependency on Western cohorts also limits applicability to populations in the Global South, where environmental exposures, health system constraints, and sociodemographic risk patterns differ markedly. 
Moreover, most models employ simple additive scoring functions, limiting their ability to capture complex interactions among risk factors. 
These limitations underscore the need for context-appropriate, low-cost, and explainable machine learning approaches tailored to early-stage CKD screening in resource-constrained environments.

Machine learning (ML)–based predictive tools, particularly those relying on easily obtainable features such as age, blood pressure, and medical history, offer a promising avenue for improving CKD risk assessment in low-resource settings. A substantial number of studies have applied ML techniques for CKD identification, and comprehensive summaries of these works are available in several review articles~\citep{sanmarchi2023predict, delrue2024application, gogoi2025machine, sabanayagam2025artificial, khan2025comprehensive}. Notably, the vast majority of ML studies are based on the UCI-2015 CKD dataset~\citep{chronic_kidney_disease_336}, which we also used for external validation. However, many of these studies include pathology-test-based markers such as serum creatinine or eGFR as input features for model development (e.g., \citep{dharmarathne2024diagnosis, islam2023chronic, pujitha2023chronic, gogoi2025chronic, bijoy2025robot, jawad2025study, nneji2025ffs}). This practice introduces a form of \emph{information leakage} and \emph{circular reasoning}, as these biomarkers are themselves the clinical criteria used to diagnose CKD; if such values are already available, an ML model becomes redundant. Moreover, pathology tests such as serum creatinine (SC) and eGFR are often inaccessible in rural or peri-urban areas and remain financially prohibitive for many individuals in the Global South. Consequently, these ML studies may be less suitable for community-based CKD screening and may have limited applicability in resource-limited populations where access to laboratory testing is constrained. While previous machine learning studies have investigated CKD prediction, none have focused specifically on community-based early-stage CKD screening while integrating extensive feature selection, systematic comparison with existing screening tools, and external validation across multiple datasets.

Building on these observations, this study proposes a machine learning–based framework for community-level early-stage CKD screening that emphasizes clinically accessible features, systematic feature selection, and rigorous validation across multiple datasets. This study makes several distinct contributions to the CKD screening literature:
\begin{itemize}
\item It develops a community-based early-stage CKD screening framework using population-level data from Bangladesh, focusing specifically on CKD stages 1–3, where early intervention is most beneficial.

\item It systematically investigates the predictive utility of clinically meaningful feature groups (socio-demographic, lifestyle and habit, medical history, clinical examination, and pathology tests) and their combinations, reflecting their impact on CKD screening.

\item It employs a rigorous feature selection framework using ten complementary methods to identify stable and clinically meaningful predictors.

\item It derives and evaluates optimized feature subsets from (i) all available variables and (ii) variables excluding pathology tests, thereby identifying minimal yet effective predictor sets suitable for community-level screening in low-resource settings.

\item It performs systematic benchmarking against established CKD screening tools on the primary dataset using machine learning models based on the derived feature subsets, demonstrating improved predictive performance.

\item It conducts cross-dataset external validation by training models on the primary dataset using the derived feature subsets and evaluating them on multiple independent datasets, demonstrating the robustness and generalizability of these feature subsets across diverse populations.
\end{itemize}

\section{Related work}
Recent advances in machine learning (ML) have led to a growing body of research on predicting chronic kidney disease (CKD) using structured health data. Several recent surveys summarize the application of artificial intelligence techniques in CKD-related tasks, including disease detection, risk prediction, prognosis, and monitoring~\citep{sanmarchi2023predict, delrue2024application, gogoi2025machine, sabanayagam2025artificial, khan2025comprehensive}. Collectively, these reviews highlight the potential of ML to support CKD research and clinical decision-making. At the same time, they identify recurring limitations in the existing literature, including limited dataset diversity, insufficient focus on early-stage CKD detection, and challenges in translating high-performing models into clinically meaningful and practically deployable screening tools. Table~\ref{table:studies} summarizes representative recent ML-based CKD prediction studies most relevant to the scope of this work. The comparison highlights differences across studies in terms of research objectives (e.g., community-level screening, early-stage detection, or applications in low-resource settings), the types of features considered (including socio-demographic, lifestyle, clinical examination, and pathology variables), and the extent to which models rely on pathology-test or gold-standard clinical biomarkers such as serum creatinine or eGFR. It also contrasts key methodological aspects, including feature selection strategies, explainability techniques, and validation protocols. This structured comparison helps clarify how the present work differs from prior studies and motivates the methodological choices adopted in this study.

\begin{table}[!ht]
    \centering
    \caption{Summary of recent ML-based studies on CKD prediction}
    \label{table:studies}
    \setlength{\tabcolsep}{4pt}
    \begin{tabular}{@{\extracolsep{4pt}}lcl ccc ccccccc cccc}
        \hline
          &  & & \multicolumn{3}{c}{Objective} & \multicolumn{7}{c}{Considered features} &   \multicolumn{4}{c}{Approach}
        \\ 
        \cline{4-6} \cline{7-13}\cline{14-17}
        \multirow[b]{1}{*}{Study}& \multirow[b]{1}{*}{Year}&\multirow[b]{1}{*}{Dataset}  & 
        \rotatebox{90}{Community-level} & \rotatebox{90}{Early-stage screening} & \rotatebox{90}{Low-resource settings} &
        \rotatebox{90}{Socio-demographic} & \rotatebox{90}{Lifestyle and habit} & \rotatebox{90}{Medical history} & \rotatebox{90}{Clinical exam} & \rotatebox{90}{Pathology test}& \rotatebox{90}{Total} & \rotatebox{90}{Used SC or eGFR} & \rotatebox{90}{FS among all features} & \rotatebox{90}{\makecell[c]{FS excluding pathology}} & \rotatebox{90}{Explainability}& \rotatebox{90}{\makecell[c]{External validation}}\\
        \hline  
        \hline
        \citep{debal2022chronic} & 2022 & SPH & \ding{55} & $\sim$ & \ding{55} & 2 & 0 & 3 & 3 & 10 & 18 & \ding{51}
        & \ding{51} & \ding{55} & \ding{55} & \ding{55}\\

        \citep{zheng2024interpretable} & 2023 & TH & \ding{55} & $\sim$ & \ding{55} & 2 & 2 & 5 & 3 & 3 & 16 & \ding{51}
        & \ding{51} & \ding{55} & \ding{51} & \ding{55}\\

        \citep{ghosh2024investigation} & 2024 & TH & \ding{55} & $\sim$ & \ding{55} & 2 & 2 & 9 & 3 & 5 & 22 & \ding{51}
        & \ding{55} & \ding{55} & \ding{51} & \ding{55}\\

        \citep{khalil2025early} & 2025 & Bab Al-Rayan & \ding{55} & -- & \ding{55} & 1 & 0 & 2 & 0 & 7 & 10 & \ding{51}
        & \ding{51} & \ding{55} & \ding{55} & \ding{55}\\

        \citep{iftikhar2023comparative} & 2023 & Buner Medical & \ding{55} & -- & \ding{55} & 2 & 0 & 0 & 1 & 17 & 20 & \ding{55}
        & \ding{55} & \ding{55} & \ding{55} & \ding{55}\\

        \citep{iftikhar2025clinical} & 2025 & Buner Medical & \ding{55} & -- & \ding{55} & 2 & 0 & 0 & 1 & 12 & 15 & \ding{55}
        & \ding{51} & \ding{55} & \ding{51} & \ding{55}\\

        \citep{iftikhar2026intelligent} & 2026 & Buner Medical & \ding{55} & -- & \ding{55} & 2 & 0 & 0 & 1 & 17 & 20 & \ding{55}
        & \ding{55} & \ding{55} & \ding{55} & \ding{55}\\

        \citep{dharmarathne2024diagnosis} & 2024 & UCI-2015 & \ding{55} & -- & \ding{55} & 1 & 0 & 2 & 3 & 4 & 10 & \ding{51}
        & \ding{55} & \ding{55} & \ding{51} & \ding{55} \\

        \citep{islam2023chronic} & 2023 & UCI-2015 & \ding{55} & -- & \ding{55} & 0 & 0 & 0 & 3 & 5  & 8 & \ding{51}
        & \ding{51} & \ding{55} & \ding{55} & \ding{55} \\

        \citep{pujitha2023chronic} & 2023 & UCI-2015 & \ding{55} & -- & \ding{55} & 1 & 0 & 4 & 3 & 10 & 18 & \ding{51}
        & \ding{51} & \ding{55} & \ding{55} & \ding{55} \\

        \citep{gogoi2025chronic} & 2025 & UCI-2015 & \ding{55} & -- & \ding{55} & 1 & 1 & 4 & 4 & 13 & 23 & \ding{51}
        & \ding{51} & \ding{55} & \ding{51} & \ding{55} \\

        \citep{bijoy2025robot} & 2025 & UCI-2015 & \ding{55} & -- & \ding{55} & 1 & 1 & 4 & 4 & 14 & 24 & \ding{51}
        & \ding{51} & \ding{55} & \ding{51} & \ding{55} \\

        \citep{jawad2025study} & 2025 & UCI-2015 & \ding{55} & -- & \ding{55} & 1 & 0 & 2 & 1 & 2 & 6 & \ding{51}
        & \ding{55} & \ding{55} & \ding{51} & \ding{55} \\
        
        \citep{nneji2025ffs} & 2025 & UCI-2015 & \ding{55} & -- & \ding{55} & 1 & 1 & 4 & 3 & 13 & 22 & \ding{51}
        & \ding{51} & \ding{55} & \ding{51} & \ding{55}\\
        
        \citep{metherall2025machine} & 2025 &  UCI-2015 & \ding{55} & -- & \ding{55} & 1 & 1 & 4 & 4 & 14 & 24 & \ding{55}
        & \ding{55} & \ding{55} & \ding{55} & \ding{55}\\

        \citep{natarajan2025using} & 2025 & UCI-2015 & \ding{55} & -- & \ding{55} & 1 & 0 & 2 & 2 & 7 & 11 & \ding{55}
        & \ding{51} & \ding{55} & \ding{55} & \ding{55}\\

        \citep{hossain2025acd} & 2025 & UCI-2015, UCI-2023 & \ding{55} & $\sim$ & \ding{55} & 1 & 0 & 4 & 4 & 14 & 23 & \ding{51}
        & \ding{51}  & \ding{55} & \ding{51} & \ding{51}\\

        \citep{rahman2024machine} & 2024 & UCI-2015, UCI-2023 & \ding{55} & $\sim$ & \ding{55} & 1 & 1 & 3 & 3 & 12 & 20 & \ding{51}
        & \ding{51} & \ding{55} & \ding{55} & \ding{51}\\
        
        \citep{md2023intelligent} & 2023 & UCI-2015, UCI-2023 & \ding{55} & $\sim$ & \ding{55} & 1 & 1 & 4 & 1 & 6 & 13 & \ding{55}
        & \ding{51} & \ding{55} & \ding{55} & \ding{51}\\
        \hline
     \multicolumn{2}{l}{This study} & TangailBD,  & \ding{51} & \ding{51} & \ding{51} & 5 & 3 & 7 & 4 & 5 & 24 & \ding{55} & \ding{51}& \ding{51}& \ding{51}& \ding{51}\\
          &     & \multicolumn{3}{l}{(TH, UCI-2015, UCI-2023)}\\
    \hline
    \multicolumn{17}{l}{FS: Feature Selection, SC: Serum Creatinine, --: unknown as CKD stages not specified,}\\
    \multicolumn{17}{l}{$\sim$: partially, as the dataset includes both early and advanced CKD stages}
    \end{tabular}
\end{table}

Most recent ML-based CKD prediction studies~\citep{dharmarathne2024diagnosis, islam2023chronic, pujitha2023chronic, gogoi2025chronic, bijoy2025robot, jawad2025study, nneji2025ffs,  metherall2025machine, natarajan2025using, zheng2024interpretable, ghosh2024investigation} rely heavily on three publicly available datasets, most notably the earlier UCI-2015 dataset~\citep{chronic_kidney_disease_336} and the more recent UCI-2023~\citep{risk_factor_prediction_of_chronic_kidney_disease_857} and TH~\citep{Al-Shamsi2018} datasets. A few studies have developed CKD prediction models using their own datasets, such as SPH~\citep{debal2022chronic}, Bab Al-Rayan~\citep{khalil2025early}, and Buner Medical~\citep{iftikhar2023comparative} datasets. However, these datasets are primarily collected in hospital settings and provide limited information on socio-demographic characteristics, lifestyle factors, and behavioral habits, which are important determinants for community-level risk assessment. In addition, these datasets offer limited support for early-stage CKD screening. For example, the UCI-2015~\citep{chronic_kidney_disease_336} dataset does not provide CKD stage information for individual subjects, while the TH dataset~\citep{Al-Shamsi2018} primarily covers patients with CKD stages 3–5 without identifying stages at the individual level. Only the UCI-2023~\citep{risk_factor_prediction_of_chronic_kidney_disease_857} dataset includes subject-level CKD stage annotations, which are necessary for stage-aware prediction tasks. Similarly, the SPH dataset used by~\citet{debal2022chronic} reports that it includes CKD stages 1–5, but detailed stage-level annotations are not clearly described. The studies used Buner Medical dataset~\citep{iftikhar2023comparative,iftikhar2025clinical,iftikhar2026intelligent} and Bab Al-Rayan dataset~\citep{khalil2025early}  also do not explicitly specify CKD stages.
As a result, these datasets have limited applicability for developing models specifically aimed at community-based early-stage CKD screening.

Furthermore, the majority of existing studies~\citep{dharmarathne2024diagnosis, islam2023chronic, pujitha2023chronic, gogoi2025chronic, bijoy2025robot, jawad2025study, nneji2025ffs,  debal2022chronic, zheng2024interpretable, ghosh2024investigation, khalil2025early} incorporate pathology-test variables such as serum creatinine (SC) or estimated glomerular filtration rate (eGFR) as input features during model development. Because these biomarkers are themselves part of the clinical diagnostic criteria for CKD, models that rely on them may provide limited additional value for early risk screening. In addition, such pathology tests are often unavailable or costly in rural or peri-urban settings and may require laboratory infrastructure that is not readily accessible in many parts of the Global South. Consequently, models built around these variables may be less suitable for community-based CKD screening and may have limited practical applicability in resource-constrained environments where laboratory testing is not routinely available.

Few studies~\citep{iftikhar2023comparative,iftikhar2025clinical,iftikhar2026intelligent,metherall2025machine,natarajan2025using} avoid using SC or eGFR as predictors; however, their primary focus is not on developing screening models for low-resource settings. Consequently, they do not explicitly investigate feature subsets excluding pathology-test variables or evaluate models designed to operate without such inputs. In addition, none of these studies reports external validation, which limits the ability to assess the generalizability of the proposed models across different populations.

From a methodological perspective, several studies~\citep{islam2023chronic, pujitha2023chronic, gogoi2025chronic, bijoy2025robot, nneji2025ffs, debal2022chronic, zheng2024interpretable, khalil2025early, iftikhar2025clinical, natarajan2025using,hossain2025acd, rahman2024machine, md2023intelligent} have explored feature selection techniques to identify important predictors for CKD classification. However, most of these studies focus on selecting optimal features from the full set of available variables, which often include pathology-test biomarkers such as SC or eGFR. While feature selection can improve model performance and reduce dimensionality, existing studies generally do not explicitly investigate feature subsets that exclude pathology-test variables. This distinction is particularly important for community-level screening in low-resource environments, where such laboratory measurements may not be readily available. As a result, prior work provides limited insight into whether reliable CKD prediction can be achieved using only clinically accessible features.

Several studies (e.g.,~\citep{dharmarathne2024diagnosis, gogoi2025chronic, bijoy2025robot, jawad2025study, nneji2025ffs, zheng2024interpretable, ghosh2024investigation, iftikhar2025clinical,hossain2025acd}) have also increasingly incorporated explainable artificial intelligence (XAI) techniques to improve model interpretability. Methods such as SHAP or feature importance analysis have been used to identify influential predictors and provide insights into model behavior. While these approaches contribute to transparency and clinical interpretability, they apply explainability primarily as a post hoc analysis without linking it to the broader objective of developing screening models for community-level deployment.

Another important methodological aspect concerns model validation. Only a few studies~\citep{hossain2025acd, rahman2024machine, md2023intelligent} attempt cross-dataset validation by training models on the UCI-2015 dataset and testing them on the UCI-2023 dataset. However, this approach still inherits the limitations associated with both datasets, including their hospital-based nature and dependence on pathology-test variables. In addition, some of these studies~\citep{hossain2025acd, rahman2024machine} still incorporate SC or eGFR as input features, which reduces their applicability for screening in resource-constrained environments. Consequently, rigorous external validation of models designed specifically for community-level CKD screening—particularly those relying on non-pathology features—remains largely unexplored.

Overall, the existing ML-based CKD prediction literature shows several limitations for community-level early-stage screening. Most studies rely on hospital-based datasets that lack comprehensive socio-demographic and lifestyle information relevant for population-level risk assessment. Many models also depend on pathology-test variables such as serum creatinine or eGFR, which are themselves diagnostic markers and may therefore provide limited additional value for predictive modeling; moreover, these tests may be unavailable in low-resource settings. While some studies apply feature selection and explainability techniques, they rarely investigate predictor subsets that exclude pathology-test variables. In addition, rigorous external validation across independent populations remains limited. These gaps highlight the need for ML frameworks that rely on clinically accessible features and are specifically designed and validated for community-based early-stage CKD screening.

\section{Methodology}
\begin{figure}[!ht]
    \centering
    \includegraphics[width=1\linewidth]{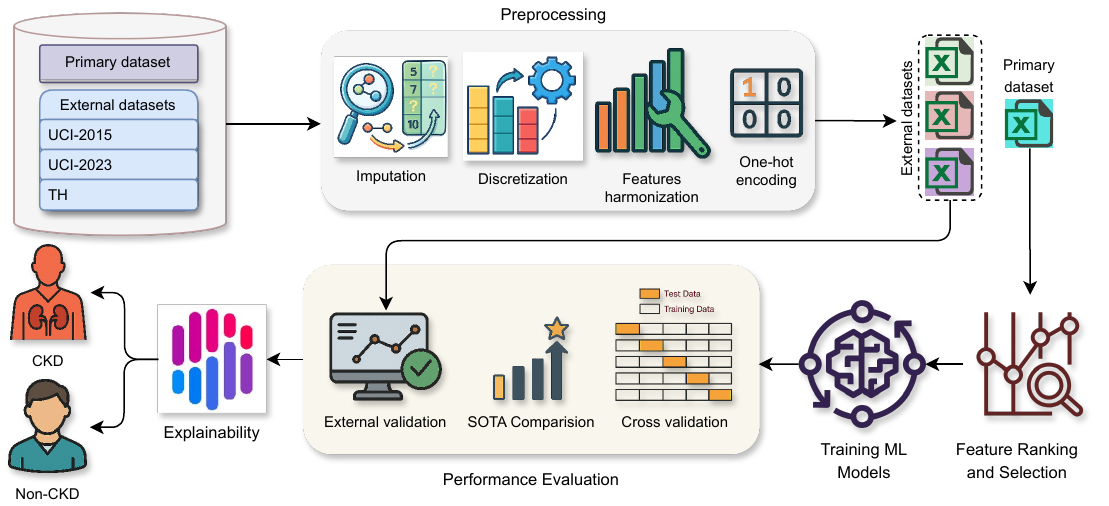}
    \caption{A schematic overview of our methodology}
    \label{fig:method}
\end{figure}

Figure~\ref{fig:method} provides an overview of the methodological pipeline used in this study. The workflow consists of several stages: 
(i) dataset preprocessing, including imputation, discretization, feature harmonization, and one-hot encoding to ensure consistency across the primary dataset and the external datasets; 
(ii) feature ranking and selection performed exclusively on the primary dataset (TangailBD) using statistical, regularization-based, and wrapper-based techniques to identify informative predictor subsets suitable for community-level early-stage CKD screening in low-resource settings; 
(iii) using these derived feature subsets, multiple machine learning classifiers are trained and evaluated on the primary dataset together with hyperparameter optimization conducted within a stratified nested cross-validation framework, where model performance was estimated using an outer stratified 10-fold cross-validation while hyperparameters were tuned within each training fold using an inner 5-fold cross-validation; 
(iv) the best-performing models are further assessed through benchmarking against established CKD screening tools on the primary dataset using the same derived feature subsets; 
(v) external validation by applying the best-performing models trained on the primary dataset to three independent datasets (UCI-2015, UCI-2023, and TH) to assess robustness and generalizability of the derived feature subsets across different populations; and 
(vi) model explainability using SHAP (SHapley Additive exPlanations) to interpret both global feature importance and individual-level predictions.

\subsection{Datasets}
The dataset used in this study originates from a community-based CKD screening conducted in the Mirzapur sub-district of Tangail, Bangladesh, a rural and peri-urban region covered by the Mirzapur demographic surveillance system (DSS). Adults aged $\geq 18$ years with at least five years of residency were selected using age-stratified random sampling, yielding 928 enrolled participants, of whom 872 completed all study procedures. Following home-based enrolment, trained community health workers obtained written informed consent, performed interviews and physical examinations, and referred participants to a hospital laboratory for the first assessment, during which blood and urine samples were collected to measure serum creatinine, fasting glucose, lipid profile, hemoglobin, and the urine albumin-to-creatinine ratio (ACR). Participants exhibiting an estimated glomerular filtration rate (eGFR) $<60$~mL/min/1.73\,m$^{2}$ and/or ACR $\geq30$~mg/g were invited for a second assessment after a 3-month interval to confirm persistent abnormalities. CKD diagnosis was made using the CKD-EPI equation to estimate eGFR and stage following the National Kidney Foundation Kidney Disease Outcomes Quality Initiative (NKF/KDOQI) guidelines~\citep{inker2014kdoqi}, requiring reduced eGFR and/or albuminuria sustained for at least three months. The study received ethics approval from the ICDDR,B Research Review Committee and Ethical Review Committee (registration no. 19081), and all participants provided written informed consent. Further details on the data collection process are provided in~\citep{sarker2021community}.

In this study, we focused exclusively on participants with CKD stages 1 to 3, as these stages represent the early phase of kidney impairment when symptoms are typically absent and community-based screening can still play a critical preventive role. Participants in stages 4 (n=5) and 5 (n=1) were excluded because these advanced stages are comparatively rare in the dataset and are usually accompanied by clinically evident symptoms, making them less relevant to early screening. After removing records with missing values, the final analytic dataset comprised 284 participants, including 112 CKD (stage 1: n = 25; stage 2: n = 51; stage 3: n = 36), and 172 non-CKD controls. Table~\ref{tab:dataset_features} presents the descriptive characteristics of the 24 study variables, all of which were transformed into categorical variables for analysis. These variables span multiple groups, including socio-demographic (SD) factors (age group, gender, literacy status, occupation, and marital status), lifestyle and habit (LH) variables (sleep duration, tobacco smoking, and smokeless tobacco use), medical history (MH) (hypertension, diabetes, heart disease, stroke, and family histories of major conditions), and physical or clinical examination (CE) (body mass index (BMI) category, abdominal obesity, undernutrition status, and anemia), and pathology tests (Path) (presence of urinary red blood cells, low serum albumin, hypercholesterolemia, low HDL-cholesterol, and hypertriglyceridemia). These variables collectively represent the multidimensional risk profile relevant to early-stage CKD screening. 

\begin{table}[!htp]
    \centering
        \caption{Primary dataset (TangailBD) description}
    \label{tab:dataset_features}
    \resizebox{\textwidth}{!}{
    \begin{tabular}{c llllcc}
    \hline
    Group & Feature & Description & Value & \makecell[c]{CKD\\[-4pt]\textit{n}=112 (\%)} & \makecell{Non-CKD\\[-4pt]\textit{n}=172 (\%)}\\
    \hline\hline
    
    \multirow{13}{*}{\rotatebox{90}{Socio-demographic (SD)}}
    & Age &  Age in years & 18-30y & 3 (2.7) & 52 (30.2)\\
    & & &31-39y & 12 (10.7) & 22 (12.8)\\
    & & &40-49y & 15 (13.4) & 47 (27.3)\\
    & & &50-60y & 33 (29.5) & 51 (29.6)\\
    & & &60+y & 49 (43.7) & 0 (0)\\
    & Gender & Male or Female & Female & 73 (65.2) & 95 (55.2)\\
    & Illiterate & Without formal education & Yes & 49 (43.7) & 55 (32)\\
    & Occupation & Occupation of participant & Farmer & 8 (7.1) & 12 (7)\\
    & & & Housewife & 67 (59.8) & 80 (46.5)\\
    & & & Others & 37 (33)  & 80 (46.5)\\
    & Marital status & Participant marital status & Married & 81 (72.3) & 142 (82.6)\\
    & & & Widowed & 28 (25) & 8 (4.6)\\
    & & & Others &  3 (2.7) & 22 (12.8)\\ 
    \hline
    
    \multirow{3}{*}{\rotatebox{90}{\makecell{Lifestyle \&\\[-4pt]habit (LH)}}}
    & Sleeping duration & $<$7 hours daily sleep (Yes) & Yes & 46 (41.1) & 35 (20.3)\\
    &Tobacco smoker & Currently smoke any form of tobacco & Yes &  13 (11.6) & 37 (21.5)\\
    & Smokeless tobacco & Currently consumed any form of smokeless tobacco & Yes & 40 (35.7) & 35 (20.3)\\
    \hline

    \multirow{7}{*}{\rotatebox{90}{Medical history (MH)}}
    &Hypertension & History of or screening-detected hypertension & Yes & 75 (67) & 20 (11.6)\\
    &Diabetes & History of or screening-detected diabetes & Yes & 30 (26.8) & 24 (13.9)\\
    &Heart disease & History of heart disease & Yes & 11 (9.8) & 6 (3.5)\\
    &Stroke & History of stroke &  Yes & 10 (8.9) & 5 (2.9)\\
    &Family diabetes & Family history of diabetes &  Yes & 28 (25) & 37 (21.5)\\
    &Family hypertension & Family history of hypertension & Yes & 43 (38.4) & 61 (35.5)\\
    &Family CKD & Family history of CKD & Yes & 12 (10.7) & 14 (8.1)\\
    \hline
    
    \multirow{7}{*}{\rotatebox{90}{Clinical examination (CE)}}
    &BMI & Body mass index, underweight ($<$18.5), & Underweight & 10 (8.9) & 23 (13.4)\\
    & & \hspace{2cm}{normal (18.5$\ge$BMI$<$25),} & Normal & 50 (44.6) & 93 (54.1)\\
    & & \hspace{2cm}{overweight (25$\ge$BMI$<$30),} & Overweight & 8 (7.1) & 13 (7.6)\\
    & & \hspace{2cm}{obese ($\ge$30)} & Obese & 44 (39.3) & 43 (25)\\   
    &Abdominal obesity & Waist circumference in cm, men$\ge$94cm, women$\ge$80cm & Yes & 59 (52.7) & 61 (35.5)\\
    &Undernutrition & Mid-upper arm circumference, men$<$25cm, women$<$24cm & Yes & 20 (17.9) & 36 (20.9)\\
    &Anemia & Blood hemoglobin, men$<$13g/dL, women$<$12g/dL & Yes & 41 (36.6) & 33 (19.2)\\
    \hline

    \multirow{5}{*}{\rotatebox{90}{Pathology tests (Path)}}
    & RBC & Presence of red blood cells in urine & Yes & 20 (17.9) & 0 (0)\\
    & Low serum albumin & Serum albumin$<$3.5 g/dL & Yes & 12 (10.7) & 4 (2.3)\\
    & Hypercholesterolaemia & Serum cholesterol$>$200 mg/dL & Yes & 28 (25) & 27 (15.7)\\
    & Low HDL-cholesterol & Serum HDL-cholesterol$<$40mg/dL & Yes & 32 (28.6) & 63 (36.6)\\
    & Hypertriglyceridemia & Serum triglyceride$>$150mg/dL & Yes & 48 (42.8) & 56 (32.6)\\
    \hline
    \end{tabular}
    }
\end{table}

For external validation, we incorporated three publicly available datasets (summarized in Table~\ref{table:datasets}). 
\begin{table}[!ht]
    \centering
    \caption{Summary of CKD datasets used in this study}
    \label{table:datasets}
    \setlength{\tabcolsep}{5pt}
    \begin{tabular}{ll rrr c cccccc c}
        \hline
        \multirow{2}{*}{Dataset}  & \multirow{2}{*}{\makecell{Country}} & \multicolumn{3}{c}{Sample size} & \multirow{2}{*}{\makecell[c]{CKD\\[-4pt]Stage}} & \multicolumn{7}{c}{Features} \\ \cline{3-5}\cline{7-13}
        & & CKD & Non-CKD & Total & & SD & LH & MH & CE & Path & Path*& Total \\
        \hline
        \hline
         TangailBD & Bangladesh & 112 & 172 & 284 & 1--3 & 5 & 3 & 7 & 4 & 5 & 5 & 29 \\
        UCI-2015 \citep{chronic_kidney_disease_336} & India & 250 & 150 & 400 & -- & 1 & 1 & 3 & 5 & 12 & 2 & 24 \\
        UCI-2023 \citep{risk_factor_prediction_of_chronic_kidney_disease_857} & Bangladesh & 128 & 72 & 200 & 1--5 & 1 & 1 & 4 & 6 & 12 & 3 & 27\\
        
        TH \citep{Al-Shamsi2018} & UAE & 56 & -- & 56 & 3--5 & 2 & 1 & 10 & 3 & 3 & 2 & 21\\\hline
        \multicolumn{13}{l}{{\small*Gold-standard pathology tests for CKD diagnosis: serum creatinine, urinary albumin, urinary creatinine, eGFR, and uACR}}
    \end{tabular}
\end{table}
The UCI-2015 CKD dataset \citep{chronic_kidney_disease_336}, collected from a hospital in India, is the most widely used benchmark dataset for machine learning research on CKD detection. It contains clinical records from 400 patients admitted over a two-month period, comprising 250 CKD cases and 150 non-CKD controls. The dataset includes 24 features spanning multiple groups: one socio-demographic variable, one lifestyle and habit-related variable, four medical history variables, four clinical examination measures, and fourteen pathology test results, including two standard pathology tests for CKD diagnosis, such as urinary albumin and serum creatinine. Notably, CKD stages are not reported in this dataset.

Another dataset used for external validation is the UCI-2023 CKD dataset \citep{risk_factor_prediction_of_chronic_kidney_disease_857}, collected at Enam Medical College in Savar, Bangladesh. It contains records from 200 patients, including 128 confirmed CKD cases and 72 non-CKD controls, with CKD stages 1–5 explicitly annotated. The dataset comprises 27 features spanning several groups: one socio-demographic attribute, one lifestyle and habit-related variable, four medical history indicators, six clinical examination measurements, and fifteen pathology test results, including three standard pathology tests for CKD diagnosis, such as urinary albumin, serum creatinine, and eGFR.

The third dataset is the TH CKD dataset \citep{Al-Shamsi2018}, derived from hospital records at Tawam Hospital in Al-Ain City, Abu Dhabi, and is one of the few Middle Eastern clinical datasets used for CKD risk modeling. It includes 491 adult patients, of whom 56 were identified with moderate-to-advanced CKD (stages 3--5). The remaining 435 participants were labeled as the ``non-CKD'' category. However, because early-stage CKD (stages 1--2) was not distinguished from healthy controls, this negative class does not constitute a clinically reliable non-CKD population. Consequently, these samples were excluded from cross-validation in our study, and only the confirmed CKD cases were used for external evaluation. The dataset contains 21 features spanning multiple groups: two socio-demographic variables, one lifestyle and habit-related factor, ten medical history attributes, three clinical examination measures, and five pathology test indicators, including two standard pathology tests for CKD diagnosis, such as serum creatinine and eGFR.

\subsection{Data Preprocessing}
Among the four datasets incorporated into this study, the UCI-2015 CKD dataset~\citep{chronic_kidney_disease_336} contained missing values requiring systematic preprocessing. Missing numeric features were addressed using the multivariate imputation by chained equations (MICE) algorithm, a widely adopted multiple-imputation framework that iteratively models each variable with missing data as a function of the other variables in the dataset~\citep{vanBuuren2011MICE}. Rather than replacing missing entries with a single deterministic value, MICE generates multiple plausible imputations, thereby preserving the underlying variability and reducing bias associated with single-imputation techniques. For categorical variables, missing entries were imputed using the most-frequent strategy, an approach demonstrated to be effective for maintaining distributional integrity in classification-focused machine learning tasks~\citep{adhikari2022comprehensive}.

To ensure consistency, shared features across external and primary datasets were identified, harmonized, and renamed. Continuous variables were discretized into categorical, where appropriate, following clinically grounded categories defined in the original cohort study (outlined in Table~\ref{tab:dataset_features}). Using these clinically grounded thresholds makes the resulting models more interpretable and better aligned with how risk factors are assessed in community health settings. Additionally, discretization helps reduce sensitivity to measurement noise and can improve model robustness when working with relatively small datasets. This discretization process aligned heterogeneous sources, enabling model training on the primary dataset and evaluation on external datasets.

Finally, all four datasets were transformed using one-hot encoding to convert nominal categorical variables into a numerical format compatible with machine learning algorithms~\citep{pedregosa2011scikit}. This approach avoids spurious ordinal assumptions, improves interpretability, and ensures consistency across the diverse classification models evaluated in this study. 

\subsection{Feature Ranking and Selection}

In this study, all variables were grouped into five conceptually coherent feature groups: socio-demographic (SD), lifestyle and habit (LH), medical history (MH), clinical examination (CE), and pathological tests (Path). These groups were defined based on clinical relevance and their potential utility for early CKD detection. The grouping also reflects their relative availability in community or low-resource settings. For example, pathology test features are typically the least accessible and least affordable in rural or resource-constrained environments, whereas CE features can be obtained by trained personnel using inexpensive portable devices. In contrast, SD, LH, and MH features can be collected through structured interviews without specialized equipment. To further examine the combined predictive utility of these groups, we also evaluated three merged feature groups: SD-LH, SD-LH-CE, and SD-LH-MH-CE.

Beyond group-wise evaluation, we employed a diverse set of feature selection (FS) techniques to identify the most informative subset of predictors across all features (denoted as set S1) and across all features except the pathology tests group (denoted as set S2). Ten complementary FS methods were applied, including a statistical filter technique such as Mann–Whitney \emph{U} (MWU) test~\citep{mann1947test}, a regularization-based method such as the least absolute shrinkage and selection operator (LASSO)~\citep{tibshirani1996lasso}, and wrapper-based recursive approaches such as recursive feature elimination with cross-validation (RFECV)~\citep{guyon2002rfe} using a variety of ML estimators, including logistic regression (LR)~\citep{hosmer2013applied}, decision tree (DT)~\citep{breiman1984cart}, random forest (RF)~\citep{breiman2001rf}, gradient boosting (GB)~\citep{friedman2001gbm}, adaptive boosting (AB)~\citep{freund1997adaboost}, extra trees (ET)~\citep{geurts2006extratrees}, extreme gradient boosting (XGB)~\citep{chen2016xgboost}, and CatBoost (CB)~\citep{dorogush2018catboost}. 

Following the application of these ten feature selection methods, we derived two consolidated feature subsets: (i) an \textit{intersection set}, representing features consistently selected across all ten FS techniques, and (ii) a \textit{union set}, comprising the complete set of unique features identified by any of the FS methods. These aggregated sets allow robust comparison of model performance under conservative (intersection) and inclusive (union) feature selection paradigms.

Feature ranking and selection were performed on the full primary dataset (TangailBD) prior to model training to identify stable predictor subsets suitable for community-based screening. Ten complementary feature selection techniques were applied to evaluate the importance of candidate variables and derive optimized feature subsets. Two subsets were constructed: S1, selected from all available variables, and S2, selected from variables excluding pathology-test features. These subsets were subsequently used for model development and evaluation. Because the goal of this step was to identify clinically meaningful predictors rather than tune model parameters, feature selection was performed once on the full dataset. The predictive performance of models built on these subsets was then assessed using nested cross-validation on the primary dataset and external validation on independent datasets.  

\subsection{Machine Learning Classifiers}
\label{sec:ml-classifiers}
To develop a robust and generalizable model for early CKD prediction, we employed twelve state-of-the-art ML classifiers spanning diverse algorithmic paradigms. These classifiers were selected based on their demonstrated effectiveness in handling structured, tabular, and high-dimensional clinical data, as well as their widespread use in prior CKD detection research~\citep{sanmarchi2023predict,delrue2024application,gogoi2025machine}. The models encompass linear, nonlinear, ensemble-based, and neural network–based learning strategies, thereby ensuring comprehensive exploration of the feature space.

The set includes a linear model such as logistic regression (LR)~\citep{hosmer2013applied}, a tree-based classifier such as decision tree (DT)~\citep{breiman1984cart}, a distance-based model such as $k$-nearest neighbours (KNN)~\citep{cover1967knn}, and a kernel-based algorithm such as support vector machine (SVM)~\citep{cortes1995svm}. To capture variance via bootstrap aggregation (bagging), we incorporated random forest (RF)~\citep{breiman2001rf} and extra trees (ET)~\citep{geurts2006extratrees}. For sequential error-correcting boosting strategies, we employed adaptive boosting (AB)~\citep{freund1997adaboost}, gradient boosting (GB)~\citep{friedman2001gbm}, extreme gradient boosting (XGB)~\citep{chen2016xgboost}, light gradient boosting (LGB)~\citep{ke2017lightgbm}, and CatBoost (CB)~\citep{dorogush2018catboost}. Finally, we included a neural network–based classifier, the multi-layer perceptron (MLP)~\citep{rumelhart1986mlp}, to evaluate nonlinear representations learned through back-propagation.
Together, these twelve classifiers provide a comprehensive benchmark for assessing CKD predictability across various modeling assumptions and complexity levels.
 
\subsection{Model Training and Evaluation}
All machine learning models were trained following a rigorous and reproducible experimental pipeline. To ensure robust estimation of generalization performance, we employed stratified 10-fold cross-validation, a widely used model evaluation strategy that mitigates overfitting by repeatedly training and testing models on different partitions of the data while preserving class distribution~\citep{kohavi1995cv}. For reproducibility, all random processes (data shuffling, model initialization, and parameter sampling) were controlled using a fixed \texttt{random\_state = 42}. To further assess model generalizability, external validation was subsequently performed using three independent datasets (UCI-2023, UCI-2015, and TH). 

Model development and evaluation were implemented using the \texttt{scikit-learn} library \citep{pedregosa2011scikit}, along with specialized optimization frameworks for advanced gradient boosting models. To identify the optimal hyperparameters for each classifier, we utilized \texttt{Optuna}, an efficient, modern hyperparameter optimization framework that employs sequential model-based optimization (SMBO) with adaptive sampling strategies to search high-dimensional parameter spaces \citep{akiba2019optuna}. Hyperparameter optimization was conducted within a nested cross-validation framework. An outer stratified 10-fold cross-validation was used to estimate model performance, while hyperparameters were tuned within each training fold using an inner 5-fold cross-validation. This design helps prevent information leakage during hyperparameter tuning and provides an unbiased estimate of generalization performance.
This systematic training and optimization procedure enabled the selection of high-performing, well-calibrated models across all feature subsets and ensured fair comparison among the twelve machine learning classifiers.

Model stability and variance were evaluated using 10-fold cross-validation. For each fold, models were trained on 90\% of the data and evaluated on the held-out fold, producing out-of-fold predictions for all samples. Performance metrics were computed for each test fold and averaged across the 10 folds to obtain overall estimates of generalization performance. To quantify uncertainty and variability, 95\% confidence intervals were estimated using stratified nonparametric bias-corrected and accelerated (BCa) bootstrap resampling (10,000 repetitions) of the pooled out-of-fold predictions. Tree-based and boosting models can exhibit variability due to their dependence on training data composition; however, cross-validation and bootstrap confidence intervals provide an empirical assessment of this variability and help ensure that reported performance reflects stable model behavior. 
To assess whether the performance difference between two models was statistically significant, we conducted a two-sided paired permutation test with 10,000 permutations using per-sample Brier loss calculated on the same evaluation samples for both models, with statistical significance assessed at $\alpha=0.05$.

Given that the dataset used in this study is imbalanced, with the CKD class representing the minority and clinically significant positive class, we employed evaluation metrics that are robust to class imbalance and that reflect clinically meaningful performance. For binary classification (CKD as the positive class), we report balanced accuracy, sensitivity (recall for CKD), area under the receiver operating characteristic curve (AUC-ROC), F1 score (for CKD), and macro-precision. The definition of those metrics is provided in \ref{app:performance:metric}. For binary classification, predicted probabilities were converted to CKD/non-CKD labels using a fixed decision threshold of 0.5, which was applied consistently across the primary dataset and all external validation datasets to ensure comparability of performance metrics. 

The best-performing model for each feature set was selected primarily based on the highest balanced accuracy, reflecting the need to account for class imbalance in CKD prediction. In cases where two or more models achieved identical balanced accuracy, we applied a secondary ranking strategy prioritizing higher sensitivity for the CKD (positive) class, followed by AUC-ROC, and subsequently the CKD-specific F1 score. This hierarchical selection process ensures that the chosen model not only performs well overall but also aligns with clinical priorities by maximizing the correct identification of CKD cases.

\subsection{Explainability}
Among the different explainability methods, SHAP (SHapley Additive exPlanations)~\citep{lundberg2017shap} has emerged as one of the most powerful due to its rigorous theoretical grounding in cooperative game theory. SHAP provides both global interpretability by quantifying overall feature importance across the population and local interpretability, allowing an examination of how specific features contribute to the prediction for an individual patient. These dual capabilities make SHAP especially suitable for clinical environments, where understanding both broad model behaviour and case-specific reasoning is essential. SHAP has also been successfully applied in previous studies on CKD prediction~\citep{gogoi2025machine}.

Motivated by its demonstrated effectiveness and theoretical strengths, this study employs SHAP to analyze the contribution of each selected feature to the model's predictions. Through this approach, we aim to bridge the gap between predictive performance and clinical interpretability, advancing the development of a transparent, explainable, and trustworthy AI-driven tool for early CKD detection.

\section{Results}

\subsection{Selected Features}\label{sec:selected features}
Table~\ref{tab:feature_ranking} summarizes the ranking of features obtained using ten complementary feature selection techniques for both S1 (all features) and S2 (all features excluding pathology tests). A small subset of features consistently appears among the highest-ranked predictors across nearly all methods, most notably \texttt{Hypertension}, \texttt{Age\_\textit{60+years}}, and the pathology-related feature \texttt{RBC}. Their consistently low rank values indicate strong and stable predictive importance, underscoring their critical role in identifying individuals at risk of CKD. Likewise, several clinically recognized risk factors—such as \texttt{Anemia}, \texttt{Diabetes}, \texttt{Daily sleep$<$7h}, and \texttt{BMI} are frequently assigned favourable ranks across multiple selection strategies, reinforcing their well-established relevance in CKD detection.

\begin{landscape}
\begin{table}[!ht]
    \centering
    \caption{Feature sets with the ranking of features (where lower rank values indicate higher predictive importance) for S1 (all features) and S2 (all features excluding the pathology test group), derived using ten diverse feature ranking and selection techniques, highlighting consistent high-impact predictors for early CKD detection.}
    \label{tab:feature_ranking}
        \setlength{\tabcolsep}{4pt}
    \begin{tabular}{@{\extracolsep{4pt}}ll |clclclclclclclc lclcl}
    \hline
      \multirow{2}{*}{Feature} &  \multirow{2}{*}{Group} & \multicolumn{16}{c}{RFECV based feature ranking} &  \multicolumn{2}{c}{\multirow{2}{*}{MWU}}&  \multicolumn{2}{c}{\multirow{2}{*}{LASSO}}\\ \cline{3-18}
         & & \multicolumn{2}{c}{LR}& \multicolumn{2}{c}{RF}& \multicolumn{2}{c}{GB}& \multicolumn{2}{c}{DT}& \multicolumn{2}{c}{AB}& \multicolumn{2}{c}{ET}& \multicolumn{2}{c}{XGB}& \multicolumn{2}{c}{CB}&  &&  &\\
         \cline{3-4}\cline{5-6}\cline{7-8}\cline{9-10}\cline{11-12}\cline{13-14}\cline{15-16}\cline{17-18}\cline{19-20}\cline{21-22}
 & & S1& S2&  S1&S2&  S1&S2&  S1&S2&  S1&S2&  S1&S2&  S1&S2&  S1&S2&  S1&S2& S1&S2\\
    \hline\hline
    Hypertension & MH & 3  &2& 1  &1& 1  &1& 1  &1& 3  &2& 1  &1& 3  &2&  1  &1& 1  &1& 3    &2\\
    Age\_\textit{60+y} & SD & 1  &1& 2  &2& 2  &2& 2  &2& 1  &1& 2  &2& 1   &1& 2  &2& 2  &2& 1 &1\\
    RBC & Path & 2  && 3  && 3  && 3  && 2  && 3  && 2  && 6  && 3  && 2  &\\  
    Anemia & CE &   && 5  &4& 4  &4&  &3&  && 6  &4& 5  &3&  4  &4& 7  &6& 10  &\\
    Diabetes & MH &  && 4  &3& 5  &3&  &4&  && 4  &3& 7   &4& 3  &3& 11  &10&  &\\
    Daily sleep $<$7h & LH &  &&  &&  &6&  &5&  &&  && 8   &6&7  && 6  &5& 5  &4\\
    Age\_\textit{18-30y} & SD &  &&  &&  &5&  &&  && 5  &5&   &&  && 4  &3& 4  &3\\
    BMI\_\textit{Obese} & CE &  &&  &&  &&  &&  &&  &&  &&   5  && 12  &11& 6  &\\

    Smokless tobacco & LH &  &&  &&  &&  &&  &&  &&   &&  && 9  &8& 9  &\\
    Age\_\textit{40-49y} & SD &  &&  &&  &&  &&  &&  &&   &&  && 10  &9& 8  &\\

 MaritialS\_\textit{Widowed} & SD &  &&  &&  &&  &&  &&  &&   &&  && 5  &4&  &\\
    Low serum albumin & Path &  &&  && 6  &&  &&  &&  && 4  &&  &&  &&  &\\
     Abdominal obesity & CE &  &&  &&  &&  &&  &&  &&  &&   && 8  &7&  &\\
       Gender & SD &  &&  &&  &&  &&  &&  &&  &&  8  &&  && 7  &\\
    Occup\_\textit{Housewife} & SD &  &&  &&  &&  &&  &&  &&  &&   &&  && 11  &5\\
    Family hypertension & MH &  &&  &&  &&  &&  &&  &&   &&9  &&  && 12  &\\
    Stroke history & MH & &&  &&  &&  &&  &&  &&  &5&   &&  &&   &\\
     BMI\_\textit{Underweight} & CE &  &&  &&  &&  &&  &&  && 6   &&  &&  &&  &\\
     BMI\_\textit{Normal} & CE &  &&  &&  &&  &&  &&  && 9  &&   &&  &&  &\\
    \hline
    \end{tabular}
\end{table}
\end{landscape}

Although some degree of variability is observed across the different methodological families, model-driven approaches (e.g., RFECV), regularization-based techniques (e.g., LASSO), and statistical tests (e.g., Mann-Whitney \emph{U} (MWU) test), the overall consensus on the top-ranked features highlights the robustness and stability of the feature selection process. Collectively, these consistently identified predictors form a strong, clinically meaningful foundation for developing reliable and generalizable CKD prediction models.

In addition to individual rankings, we derived both \textit{common} and \textit{union} feature subsets for S1 and S2. The common subset for S2 comprises \texttt{Hypertension} and \texttt{Age\_\textit{60+years}}, while the S1 common subset additionally includes the pathology-based feature \texttt{RBC}. These shared features represent the most universally selected predictors across all methods, providing compact yet highly informative feature sets for downstream model development.

\subsection{Performance of ML Models}


Table~\ref{tab:performance} summarizes the performance of all feature sets, reporting results based on the best-performing model for each set. Every feature set was evaluated independently using the twelve machine learning classifiers introduced in Section~\ref{sec:ml-classifiers}. 
\begin{table}[!htp]
\centering
\caption{Predictive performance of the best machine learning models for different feature groups, feature combinations, and feature-selection. Results are reported as mean performance across 10-fold cross-validation with 95\% confidence interval estimated using stratified nonparametric BCa bootstrap resampling of pooled fold predictions. The best-performing classifier for each feature set is reported. Bold values represent the highest performance metric in each group. \label{tab:performance}}
\renewcommand{\arraystretch}{.87}
\begin{tabular}{c llccccc}
\hline
&\multirow{2}{*}{\makecell{Feature set}} &  \multirow{2}{*}{\makecell[t l]{Best\\[-4pt]Model}} & \multicolumn{5}{c}{Performance with 95\% confidence interval (CI)}\\
\cline{4-8}
& & &  \makecell{Balanced\\[-4pt]accuracy}  & \makecell{Sensitivity\\[-4pt](CKD)} 
& AUC-ROC & \makecell{F1\\[-4pt](CKD)} & \makecell{Precision\\[-4pt](macro)} \\
\hline
\hline
& All & CB & $84.54_{\text{79.6--88.5}}$ & $79.55_{\text{69.6--85.7}}$ & $0.900_{\text{0.85--0.93}}$ & $81.22_{\text{75.2--86.2}}$ & $85.30_{\text{80.1--88.9}}$ \\

\hline
\multirow{8}{*}{\rotatebox{90}{Groups and combinations}}
&Socio-Demo (SD) & DT & $73.27_{\text{67.7--78.3}}$	& $65.23_{\text{54.5--72.3}}$	& $0.816_{\text{0.75--0.86}}$	& $67.08_{\text{60.1--73.8}}$	& $74.35_{\text{68.2--79.0}}$ \\

&Life-style/Habbit (LH) & LGB & $62.90_{\text{57.2--68.5}}$	& $53.64_{\text{43.8--61.6}}$	& $0.648_{\text{0.53--0.67}}$	& $54.09_{\text{47.2--61.9}}$	& $63.18_{\text{57.2--68.6}}$ \\

&Medical-history (MH) & DT & $78.87_{\text{73.5--83.2}}$	& $75.15_{\text{65.2--81.3}}$	& $0.805_{\text{0.73--0.85}}$	& $74.37_{\text{67.9--79.8}}$	& $78.99_{\text{73.1--83.0}}$ \\

&Clinical exam (CE) & LR & $58.77_{\text{52.9--64.6}}$	& $57.12_{\text{47.3--65.2}}$	& $0.593_{\text{0.51--0.65}}$	& $51.68_{\text{45.6--59.0}}$	& $58.59_{\text{52.8--64.0}}$ \\

&Pathology tests (Path) & LR & $64.24_{\text{58.3--69.6}}$	& $52.80_{\text{42.9--60.7}}$	& $0.702_{\text{0.60--0.73}}$	& $55.12_{\text{47.5--62.8}}$	& $66.13_{\text{58.7--70.5}}$ \\

&SD-LH & AB & $73.20_{\text{68.2--77.9}}$	& $49.92_{\text{39.3--58.0}}$	& $0.820_{\text{0.76--0.86}}$	& $63.95_{\text{55.4--72.0}}$	& $\textbf{83.53}_{\text{77.0--86.3}}$ \\

&SD-LH-MH & LR & $\textbf{82.93}_{\text{77.9--87.0}}$	& $\textbf{78.64}_{\text{68.8--84.8}}$	& $0.881_{\text{0.83--0.92}}$	& $\textbf{79.16}_{\text{73.2--84.4}}$	& $83.40_{\text{78.2--87.2}}$ \\

&SD-LH-MH-CE & CB & $81.87_{\text{76.9--86.2}}$	& $74.17_{\text{64.3--81.3}}$	& $\textbf{0.887}_{\text{0.84--0.93}}$	& $77.80_{\text{71.5--83.5}}$	& $83.21_{\text{78.2--87.3}}$ \\
\hline

\multirow{12}{*}{\rotatebox{90}{Selected across all features (S1)}}
&RFECV+LR & AB & $86.19_{\text{81.4--90.0}}$	& $84.02_{\text{75.0--89.3}}$	& $0.895_{\text{0.84--0.92}}$	& $82.91_{\text{77.6--87.8}}$	& $86.35_{\text{81.2--89.7}}$ \\

&RFECV+RF & SVM & $88.45_{\text{83.5--91.5}}$	& $76.89_{\text{67.0--83.0}}$	& $0.859_{\text{0.78--0.90}}$	& $86.24_{\text{80.2--90.7}}$	& $\textbf{93.72}_{\text{91.1--95.0}}$ \\

&RFECV+GB & GB & $88.75_{\text{84.4--92.1}}$	& $78.64_{\text{68.8--84.8}}$	& $0.920_{\text{0.85--0.94}}$	& $86.56_{\text{81.4--91.3}}$	& $92.96_{\text{89.5--94.8}}$ \\

&RFECV+DT & AB & $86.19_{\text{81.4--90.0}}$	& $84.02_{\text{75.0--89.3}}$	& $0.895_{\text{0.84--0.92}}$	& $82.91_{\text{77.6--87.8}}$	& $86.35_{\text{81.2--89.7}}$ \\

&RFECV+AB & AB & $86.19_{\text{81.4--90.0}}$	& $84.02_{\text{75.0--89.3}}$	& $0.895_{\text{0.84--0.92}}$	& $82.91_{\text{77.6--87.8}}$	& $86.35_{\text{81.2--89.7}}$	\\

&RFECV+ET & GB & $87.99_{\text{83.0--91.1}}$	& $75.98_{\text{66.1--82.1}}$	& $\textbf{0.931}_{\text{0.87--0.95}}$	& $85.76_{\text{79.6--90.2}}$	& $93.44_{\text{91.0--94.8}}$ \\

&RFECV+XGB & XGB & $88.16_{\text{83.5--91.8}}$	& $80.38_{\text{70.5--85.7}}$	& $0.915_{\text{0.86--0.95}}$	& $85.76_{\text{80.2--90.7}}$	& $91.11_{\text{86.3--93.4}}$ \\

&RFECV+CB & DT  & $\textbf{90.40}_{\text{86.1--93.5}}$	& $\textbf{84.85}_{\text{75.9--90.2}}$	& $0.907_{\text{0.86--0.94}}$	& $\textbf{88.77}_{\text{83.5--92.6}}$	& $92.14_{\text{87.9--94.7}}$ \\

&LASSO & LR & $84.99_{\text{80.1--88.9}}$	& $82.20_{\text{72.3--87.5}}$	& $0.910_{\text{0.87--0.94}}$	& $81.46_{\text{76.0--86.5}}$	& $85.56_{\text{80.1--88.7}}$ \\

&MWUtest & GB & $85.99_{\text{81.2--89.8}}$	& $79.47_{\text{69.6--85.7}}$	& $0.925_{\text{0.87--0.95}}$	& $83.09_{\text{77.3--88.0}}$	& $87.91_{\text{82.6--90.8}}$ \\

&Union & CB & $85.88_{\text{81.1--89.6}}$	& $80.45_{\text{71.4--85.7}}$	& $0.898_{\text{0.85--0.93}}$	& $82.88_{\text{77.1--87.6}}$	& $87.00_{\text{82.1--90.3}}$ \\

&Common & AB & $86.19_{\text{81.4--90.0}}$	& $84.02_{\text{75.0--89.3}}$	& $0.895_{\text{0.84--0.92}}$	& $82.91_{\text{77.6--87.8}}$	& $86.35_{\text{81.2--89.7}}$ \\
\hline

\multirow{12}{*}{\rotatebox{90}{Selected across all except pathology tests (S2)}}

&RFECV+LR & AB  & $83.50_{\text{78.5--87.6}}$	& $78.64_{\text{68.8--84.8}}$	& $0.861_{\text{0.78--0.89}}$	& $79.77_{\text{73.8--85.0}}$	& $84.21_{\text{79.0--87.9}}$ \\

&RFECV+RF & MLP  & $86.25_{\text{81.6--90.1}}$	& $77.73_{\text{68.8--83.9}}$	& $0.914_{\text{0.84--0.93}}$	& $83.14_{\text{77.7--88.5}}$	& $89.56_{\text{84.1--91.8}}$ \\

&RFECV+GB & CB  & $\textbf{89.23}_{\text{84.9--92.6}}$	& $\textbf{83.11}_{\text{74.1--88.4}}$	& $0.906_{\text{0.84--0.94}}$	& $\textbf{87.12}_{\text{82.0--91.4}}$	& $\textbf{90.92}_{\text{86.8--93.7}}$ \\

&RFECV+DT & DT  & $87.75_{\text{83.2--91.4}}$	& $79.55_{\text{70.5--85.7}}$	& $0.913_{\text{0.84--0.94}}$	& $85.11_{\text{79.8--90.1}}$	& $90.19_{\text{86.2--93.2}}$ \\

&RFECV+AB & AB & $83.50_{\text{78.5--87.6}}$	& $78.64_{\text{68.8--84.8}}$	& $0.861_{\text{0.78--0.89}}$	& $79.77_{\text{73.8--85.0}}$	& $84.21_{\text{79.0--87.9}}$ \\

&RFECV+ET & CB & $86.25_{\text{81.6--90.1}}$	& $77.73_{\text{68.8--83.9}}$	& $0.909_{\text{0.84--0.93}}$	& $83.14_{\text{77.7--88.5}}$	& $89.56_{\text{84.1--91.8}}$ \\

&RFECV+XGB & RF & $87.17_{\text{82.6--90.8}}$	& $79.55_{\text{70.5--85.7}}$	& $\textbf{0.921}_{\text{0.86--0.94}}$	& $84.24_{\text{79.0--89.3}}$	& $89.30_{\text{84.9--92.4}}$ \\

&RFECV+CB & MLP & $86.25_{\text{81.6--90.1}}$	& $77.73_{\text{68.8--83.9}}$	& $0.914_{\text{0.84--0.93}}$	& $83.14_{\text{77.7--88.5}}$	& $89.56_{\text{84.1--91.8}}$ \\

&LASSO & LR & $83.50_{\text{78.5--87.6}}$ & $78.64_{\text{68.8--84.8}}$ & $0.892_{\text{0.83--0.92}}$ & $79.77_{\text{73.8--85.0}}$ & $84.21_{\text{79.0--87.9}}$\\

&MWUtest & CB & $84.42_{\text{79.5--88.3}}$	& $78.64_{\text{69.6--84.8}}$	& $0.896_{\text{0.83--0.93}}$	& $80.84_{\text{75.1--86.1}}$	& $85.61_{\text{80.2--89.0}}$\\

&Union & LGB & $84.58_{\text{79.6--88.5}}$	& $79.62_{\text{69.6--85.7}}$	& $0.898_{\text{0.84--0.93}}$	& $80.95_{\text{75.3--86.2}}$	& $85.52_{\text{80.3--88.9}}$\\

&Common & AB  & $83.50_{\text{785--87.6}}$  &$78.64_{\text{68.8--84.8}}$ 
& $0.861_{\text{0.78--0.89}}$ & $79.77_{\text{73.8--85.0}}$ & $84.21_{\text{79.0--87.9}}$ \\
\hline
 \end{tabular}
\end{table}

Overall, the results in Table~\ref{tab:performance} highlight that models trained on the selected features from the full feature set (S1) achieve the strongest predictive performance. The highest balanced accuracy of 90.40\% (95\% CI 86.1\%--93.5\%) is obtained using a DT model applied to the RFECV with CatBoost selected S1 feature subset (Table~\ref{tab:best-s1-s2}), achieving the highest sensitivity of 84.85\% (95\% CI 75.9\%--90.2\%) and F1 score of 88.77\% (95\% CI 83.5\%--92.6\%). Among the individual feature groups, the \textit{Medical History} (MH) and \textit{Socio-Demographic} (SD) sets demonstrate competitive predictive ability, with balanced accuracies of 78.87\% (95\% CI 73.5\%--83.2\%) and 73.27 (95\% CI 67.7\%--78.3\%)\%, respectively. Notably, the combined SD-LH-MH set outperforms all individuals and combined groups, achieving a balanced accuracy of 82.93\% (95\% CI 77.9\%--87.0\%), which is higher than when clinical examination features are additionally included (i.e., SD-LH-MH-CE).
\begin{table}[!ht]
    \centering
     \caption{Best-performing feature sets identified for S1 (all features) and S2 (all features excluding pathology tests), based on the highest balanced accuracy achieved across all models.}
    \label{tab:best-s1-s2}
    \resizebox{\textwidth}{!}{
    \begin{tabular}{c llcc}
    \hline
        Best set & Features & Description & Selection approach & Best model\\
       \hline\hline
         S1 & \makecell[l]{Hypertension, Age\_\textit{60+y}, Diabetes, Anemia,\\ BMI\_\textit{Obese}, RBC, Daily sleep $<$7h,\\ Gender, Family hypertension} & \makecell[l]{Best performing feature set among\\all features \textit{including} pathology tests} & RFECV+CB & DT \\
         \hline
       S2 & \makecell[l]{Hypertension, Age\_\textit{60+y}, Anemia, Diabetes, \\Daily sleep $<$7h, Age\_\textit{18-30y}} &  \makecell[l]{Best performing feature set among\\all features \textit{excluding} pathology tests} & RFECV+GB & CB\\
        \hline
    \end{tabular}
}
\end{table}

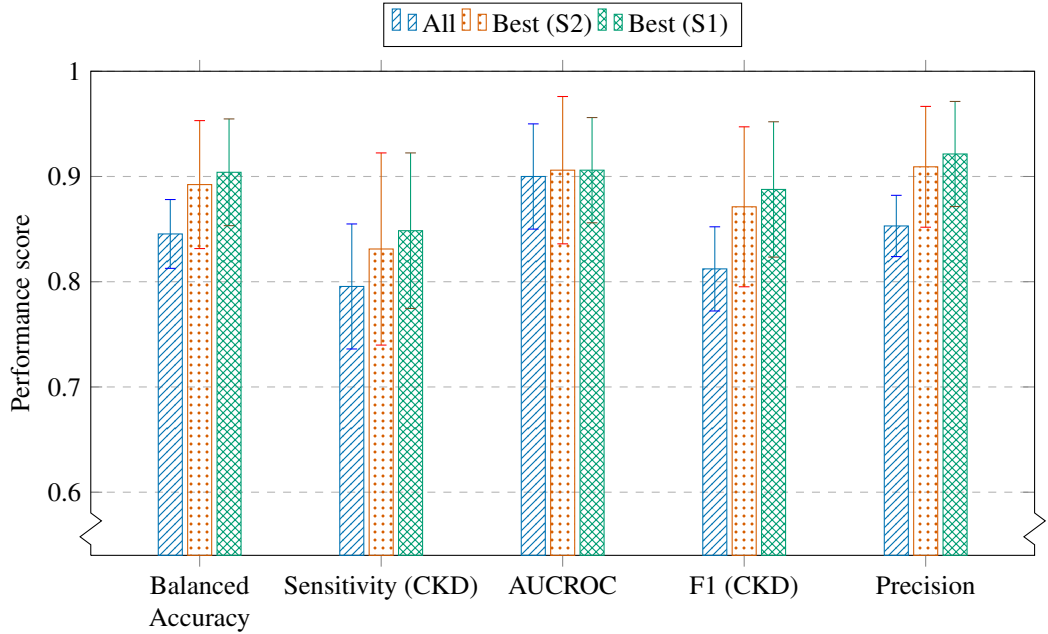
\begin{figure}[!ht]
\centering
\begin{tikzpicture}
\begin{axis}[
    ybar,
    bar width=9pt,
    width=14cm,
    height=8cm,
    enlarge x limits=0.15,
    ymin=0.54,
    ymax=1,
    axis y discontinuity=crunch,
    ylabel={Performance score},
    xtick=data,
    xticklabel style={align=center},
xticklabels={{Balanced\\[-6pt]Accuracy}, {Sensitivity (CKD)}, {AUCROC}, {F1 (CKD)},
    {Precision}
},
    xmajorgrids=false,
    ymajorgrids,
    yminorgrids,
    grid style={dashed,gray!60},
    legend style={at={(0.5,1.05)}, anchor=south, legend columns=3},
]

\definecolor{SciBlue}{rgb}{0.00,0.45,0.70}      
\definecolor{SciRed}{rgb}{0.84,0.37,0.00}       
\definecolor{SciGreen}{rgb}{0.00,0.62,0.45}     

\addplot+[pattern=north east lines, pattern color=SciBlue, draw=SciBlue,  error bars/.cd, 
          y dir=both, y explicit]
    coordinates {
        (0, 0.8454) += (0,0.0494) -= (0,0.0396)
        (1, 0.7955)  += (0,0.0995) -= (0,0.0615)
        (2, 0.9000)  += (0,0.05) -= (0,0.03)
        (3, 0.8122) += (0,0.0602) -= (0,0.0498)
        (4, 0.8530)  += (0,0.052) -= (0,0.036)
    };

\addplot+[pattern=dots, pattern color=SciRed, draw=SciRed,  error bars/.cd,
          y dir=both, y explicit]
    coordinates {
        (0, 0.8923) += (0,0.0433) -= (0,0.0337)
        (1, 0.8311) += (0,0.0901) -= (0,0.0529)
        (2, 0.9060) += (0,0.066) -= (0,0.034)
        (3, 0.8712) += (0,0.0512) -= (0,0.0428)
        (4, 0.9092) += (0,0.0412) -= (0,0.0278)
    };
    
\addplot+[pattern=crosshatch, pattern color=SciGreen, draw=SciGreen, error bars/.cd,
          y dir=both, y explicit]
    coordinates {
        (0, 0.9040) += (0,0.043) -= (0,0.031)
        (1, 0.8485) += (0,0.0895) -= (0,0.0535)
        (2, 0.9060) += (0,0.0447) -= (0,0.033)
        (3, 0.8877) += (0,0.052) -= (0,0.0383)
        (4, 0.9214) += (0,0.0424) -= (0,0.0256)
    };
\legend{All features, Best feature subset w/o pathology (S2), Best feature subset among all (S1)}
\end{axis}
\end{tikzpicture}
\caption{Performance comparison of machine learning models trained on three different feature configurations: the full feature set, the best-performing subset excluding pathology features (S2), and the best-performing subset across all feature domains (S1). Performance metrics are reported as the average across 10-fold cross-validation, and error bars represent 95\% confidence intervals estimated using stratified nonparametric BCa bootstrap resampling of pooled fold predictions from all folds, 
which may produce asymmetric intervals.
\label{fig:performance-comparison}}
\end{figure}

\begin{figure}[!ht]
    \centering
    \begin{subfigure}[b]{0.33\linewidth} 
        \centering
        \includegraphics[width=\textwidth]{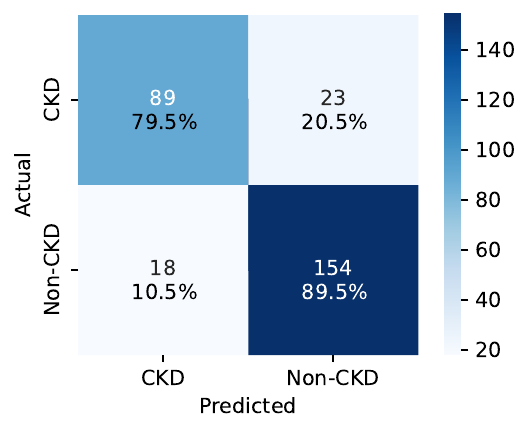}
        \caption{}
        \label{fig:cm-all}
    \end{subfigure}
    \begin{subfigure}[b]{0.33\linewidth}
        \centering
        \includegraphics[width=\textwidth]{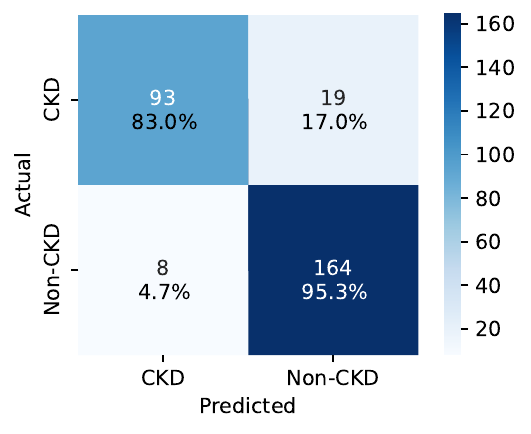}
        \caption{}
        \label{fig:cm-s2}
    \end{subfigure}
   \begin{subfigure}[b]{0.33\linewidth}
    \centering
    \includegraphics[width=\textwidth]{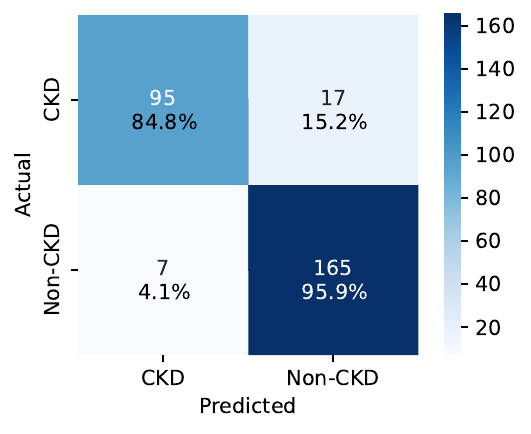}
    \caption{}
    \label{fig:cm-s1}
    \end{subfigure}
    
    \caption{Confusion matrices illustrating the prediction results for CKD and non-CKD cases using three feature configurations: (a) the full feature set, (b) the best-performing subset without pathology features, and (c) the best-performing subset including all features. Values represent both the counts and percentages of true positives, true negatives, false positives, and false negatives.}
    \label{fig:cm}
\end{figure}

Feature subsets derived through feature selection methods (for both S1 and S2) also exhibit strong predictive performance. Across both settings, RFECV-based feature selection consistently outperforms LASSO and the MWU test, highlighting the advantage of model-driven wrapper approaches. Importantly, the selected feature set from S2, which excludes pathology-related variables, achieves performance close to that of the full feature set. Using a CatBoost model on the RFECV+GB-selected S2 feature subset (Table~\ref{tab:best-s1-s2}), we obtain a balanced accuracy of 89.23\% (95\% CI 84.9\%--92.6\%), sensitivity of 83.11\% (95\% CI 74.1\%--88.4\%), F1 score of 87.12\% (95\% CI 82.0\%--91.4\%), and precision of 90.92\% (95\% CI 86.8\%--93.7\%). These values are only 1--2\% lower than the best-performing S1 models, demonstrating that CKD can be predicted with competitive accuracy using solely non-pathological, non-laboratory features. Figures~\ref{fig:performance-comparison} and~\ref{fig:cm} together illustrate the comparative performance and classification behaviour of models trained on those feature sets.
This finding underscores the feasibility of deploying CKD screening models in community-based and low-resource settings without the need for laboratory or pathology tests.


\subsection{Comparison with CKD Screening Tools}

Table~\ref{tab:sota_comparison} compares the proposed machine learning models with several established CKD screening tools, including SCORED~\citep{bang2007development}, \citet{kshirsagar2008simple} (simplified version), \citet{thakkinstian2011simple}, \citet{kearns2013predicting}, and \citet{kwon2019simple}. 
For consistency with the binary classification framework used in this study, the \citet{thakkinstian2011simple} tool, originally designed to assign individuals to four risk categories (low, intermediate-low, intermediate-high, and high), was converted into a binary prediction task by treating the \textit{low} category as non-CKD and the remaining categories as CKD. In addition, the probability output of \citet{kearns2013predicting} does not naturally scale between 0 and 1; therefore, an optimal decision threshold was determined by maximizing the F1 score on the evaluation data, resulting in a threshold value of 0.0363. A brief description of these tools is provided in~\ref{app:tools} and summarized in~\ref{tab:sota}.

As shown in Table~\ref{tab:sota_comparison}, the proposed models based on the optimized S1 and S2 feature subsets consistently outperform all existing screening tools across most evaluation metrics. To assess whether the observed improvements are statistically significant, we performed a paired permutation test based on per-sample Brier loss using the shared evaluation samples. The resulting $p$-values are reported in Table~\ref{tab:sota_comparison} for both the S1 and S2 models. The results show that the proposed models significantly outperform all existing screening tools ($p < 0.001$). These results indicate that the machine learning models derived from the proposed feature selection framework provide substantially improved predictive performance compared with traditional risk-score–based screening tools.
\begin{table}[!ht]
    \centering
\caption{Performance comparison between the proposed machine learning models and established CKD screening tools on the TangailBD dataset. S1 denotes the optimized feature subset derived from all variables, and S2 denotes the optimized subset excluding pathology-test variables. The proposed models (best S1 with Decision Tree and best S2 with CatBoost) are evaluated against traditional screening tools. Statistical significance is assessed using a paired permutation test based on per-sample Brier loss, with $p$-values indicating whether the performance differences between the proposed models and each screening tool are statistically significant.}
    \label{tab:sota_comparison}
    \resizebox{1\textwidth}{!}{
    \setlength{\tabcolsep}{3pt}
    \begin{tabular}{l lllll | cc}
    \hline
     \makecell[l]{Tool\\[-4pt]} & \multirow{2}{*}{\makecell{Balanced\\[-4pt]accuracy}} & \multirow{2}{*}{\makecell{Sensitivity\\[-4pt](CKD)}} 
& \makecell[l]{AUC-ROC\\[-4pt]} &  \multirow{2}{*}{\makecell{F1\\[-4pt](CKD)}} & \multirow{2}{*}{\makecell{Precision\\[-4pt](macro)}} & \multicolumn{2}{c}{Significance} \\ \cmidrule{7-8}
&  & & & & & S1 & S2\\
      \hline\hline
      SCORED~\citep{bang2007development}   & $76.35_{\text{70.9--81.2}}$  &$66.07_{\text{55.3--73.2}}$ 
& $0.851_{\text{0.79--0.89}}$ & $70.81_{\text{63.6--77.2}}$ & $77.98_{\text{72.4--82.6}}$ & $p<0.001$ & $p<0.001$\\ 

      \citet{kshirsagar2008simple} & $77.09_{\text{71.7--81.8}}$  &$66.96_{\text{57.1--74.1}}$ 
& $0.869_{\text{0.82--0.91}}$ & $71.77_{\text{64.6--77.9}}$ & $78.77_{\text{73.4--83.5}}$ & $p<0.001$ & $p<0.001$\\

      \citet{thakkinstian2011simple} & $81.03_{\text{75.9--85.4}}$  &$81.25_{\text{72.3--86.6}}$ 
& $0.804_{\text{0.74--0.86}}$ & $77.12_{\text{71.4--82.2}}$ & $80.13_{\text{75.2--84.5}}$ & $p<0.001$ & $p<0.001$\\

      \citet{kwon2019simple} & $77.49_{\text{72.2--82.2}}$ &$64.29_{\text{53.6--71.4}}$ 
& $0.854_{\text{0.80--0.89}}$ & $72.00_{\text{64.3--78.5}}$ & $80.71_{\text{75.1--85.2}}$ & $p<0.001$ & $p<0.001$\\
      
      \citet{kearns2013predicting} & $78.73_{\text{73.5--83.1}}$  &$83.13_{\text{74.1--88.4}}$ 
& $0.867_{\text{0.81--0.90}}$ & $74.70_{\text{69.1--79.5}}$ & $77.48_{\text{72.4--81.7}}$ & $p<0.001$ & $p<0.001$\\

      \hline
      This study (best S1 with DT) & $90.40_{\text{86.1--93.5}}$	& $84.85_{\text{75.9--90.2}}$	& $0.907_{\text{0.86--0.94}}$	& $88.77_{\text{83.5--92.6}}$	& $92.14_{\text{87.9--94.7}}$  & -- & $p=0.74$\\
      This study (best S2 with CB) & $89.23_{\text{84.9--92.6}}$	& $83.11_{\text{74.1--88.4}}$	& $0.906_{\text{0.84--0.94}}$	& $87.12_{\text{82.0--91.4}}$	& $90.92_{\text{86.8--93.7}}$ & $p=0.74$ & --\\ 
      \hline
    \end{tabular}
    }
\end{table}

Importantly, the comparison between the S1 and S2 models shows no statistically significant difference ($p = 0.74$), indicating that the reduced feature subset excluding pathology-test variables achieves performance comparable to the full feature set. This finding suggests that effective CKD screening can be achieved using minimal non-laboratory features, which are more suitable for community-level screening in resource-constrained settings.

\subsection{External Validation}

To evaluate the applicability and generalizability of the best-performing S1 and S2 feature subsets, we conducted external validation using three independent publicly available datasets: UCI-2023~\citep{chronic_kidney_disease_336}, UCI-2015~\citep{risk_factor_prediction_of_chronic_kidney_disease_857}, and TH~\citep{Al-Shamsi2018}. For each of these datasets, we first identified the set of features shared with our dataset and harmonized the naming conventions to ensure consistency across sources. Based on these harmonized features, we then constructed the S1 and S2 subsets by selecting the variables that overlapped with our optimized S1 and S2 feature sets, respectively. The resulting common, S1, and S2 feature subsets for each external dataset are summarized in Table~\ref{tab:match-feature-sets}. Notably, the TH dataset did not contain any pathology-related features that matched those in our dataset, and therefore, its S1 subset could not be constructed. This process enables a direct and fair assessment of how well the models trained on our dataset generalize when applied to diverse populations and measurement contexts.

\begin{table}[!ht]
    \centering
        \caption{Feature sets for external validation}
    \label{tab:match-feature-sets}
    \begin{tabular}{lcll}
    \hline
       Dataset  & Name & Features \\
       \hline \hline
        UCI-2023~\citep{chronic_kidney_disease_336} & Common & Hypertension, Age\_\textit{60+y}, Diabetes, Anemia, RBC\\
        & S1 subset &  Hypertension, Age\_\textit{60+y}, Diabetes, Anemia, RBC \\
        & S2 subset & Hypertension, Age\_\textit{60+y}, Diabetes, Anemia\\
        \hline
        UCI-2015~\citep{risk_factor_prediction_of_chronic_kidney_disease_857} & Common & Hypertension, Age\_\textit{60+y}, Diabetes, Anemia, RBC, Age\_\textit{18-30y},\\
        & & Age\_\textit{31-39y}, Age\_\textit{40-49y}, Age\_\textit{50-60y}\\
        & S1 subset &Hypertension, Age\_\textit{60+y}, Diabetes, Anemia, RBC\\
        & S2 subset & Hypertension, Age\_\textit{60+y}, Diabetes, Anemia, Age\_\textit{18-30y}\\
        \hline
       TH~\citep{Al-Shamsi2018} & Common & Hypertension, Age\_\textit{60+y}, Diabetes, BMI\_\textit{Obese}, Gender, Age\_\textit{31-39y},\\
       & & Age\_\textit{18-30y}, Age\_\textit{40-49y}, Age\_\textit{50-60y}, Heart disease,\\
       & & Tobacco smoker, Hypercholesterolemia, Hypertriglyceridemia\\
         & S1 subset & \textit{None common pathology tests}  \\
         & S2 subset &  Hypertension, Age\_\textit{60+y}, Diabetes, Age\_\textit{18-30y}\\
        \hline
    \end{tabular}

\end{table}

Table~\ref{tab:external-val-performance} presents the external validation performance of the machine learning models trained on our dataset and evaluated on the three independent datasets. Across all datasets, both the common and subset-based models demonstrate strong sensitivity (e.g., 80.47\%, 95\% CI 71.9\%--85.9\%, in UCI-2023) and competitive balanced accuracy (e.g., 79.33\%, 95\% CI 74.9\%--83.1\%, in UCI-2015), indicating good generalizability of the trained models to external populations. Notably, the TH dataset contains only CKD cases; therefore, only sensitivity is reported for this cohort, with the S2 subset demonstrating exceptional performance at 98.21\% (95\% CI 89.3\%--100.0\%).
\begin{table}[!ht]
\centering
\caption{External validation performance of machine learning models trained on TangailBD dataset and tested on three external datasets. 
\label{tab:external-val-performance}}

    \setlength{\tabcolsep}{4pt}
\begin{tabular}{lllccccc}
\hline
Train & Test & Feature set & B. Accuracy  & Sensitivity 
& AUC-ROC & F1 & Precision (macro) \\
\hline\hline
\multirow{8}{*}{\rotatebox{90}{TangailBD}} & 
UCI-2023~\citep{chronic_kidney_disease_336} & Common & $76.35_{\text{69.5--82.0}}$  &$80.47_{\text{71.9--85.9}}$ 
& $0.819_{\text{0.76--0.87}}$ & $82.07_{\text{76.8--86.5}}$ & $75.64_{\text{68.9--81.4}}$ \\
& & S1 subset & $76.35_{\text{69.5--82.0}}$  &$80.47_{\text{71.9--85.9}}$ & $0.844_{\text{0.79--0.89}}$ & $82.07_{\text{76.8--86.5}}$ & $75.64_{\text{68.9--81.4}}$ \\
& & S2 subset & $74.00_{\text{67.1--79.8}}$  &$75.78_{\text{66.4--82.0}}$ & $0.804_{\text{0.74--0.86}}$ & $79.18_{\text{73.4--84.0}}$ & $72.78_{\text{66.2--78.3}}$ \\
\cmidrule{2-8}

& UCI-2015~\citep{risk_factor_prediction_of_chronic_kidney_disease_857} & Common & $78.93_{\text{74.4--82.8}}$  &$77.20_{\text{71.2--81.6}}$ & $0.826_{\text{0.78--0.86}}$ & $81.78_{\text{77.8--85.2}}$ & $77.46_{\text{73.1--81.2}}$ \\
& & S1 subset & $79.33_{\text{74.9--83.1}}$  &$78.00_{\text{72.0--82.4}}$ & $0.881_{\text{0.84--0.91}}$ & $82.28_{\text{78.3--85.7}}$ & $77.90_{\text{73.7--81.6}}$ \\
& & S2 subset & $77.13_{\text{72.6--80.9}}$  &$73.60_{\text{67.2--78.2}}$ & $0.842_{\text{0.79--0.88}}$ & $79.48_{\text{75.2--83.1}}$ & $75.55_{\text{71.3--79.2}}$ \\
\cmidrule{2-8}

& TH~\citep{Al-Shamsi2018} & Common & --  &$96.43_{\text{85.6--98.2}}$ & -- & -- & -- \\
& & S2 subset & --  &$98.21_{\text{89.3--100}}$ & -- & -- & -- \\
\hline

\multicolumn{8}{l}{{B. Accuracy - Balanced Accuracy}}

\end{tabular}
\end{table}


\subsection{Model Interpretation and Key Predictors}

\begin{figure}[!ht]
    \centering
    \includegraphics[width=0.75\linewidth]{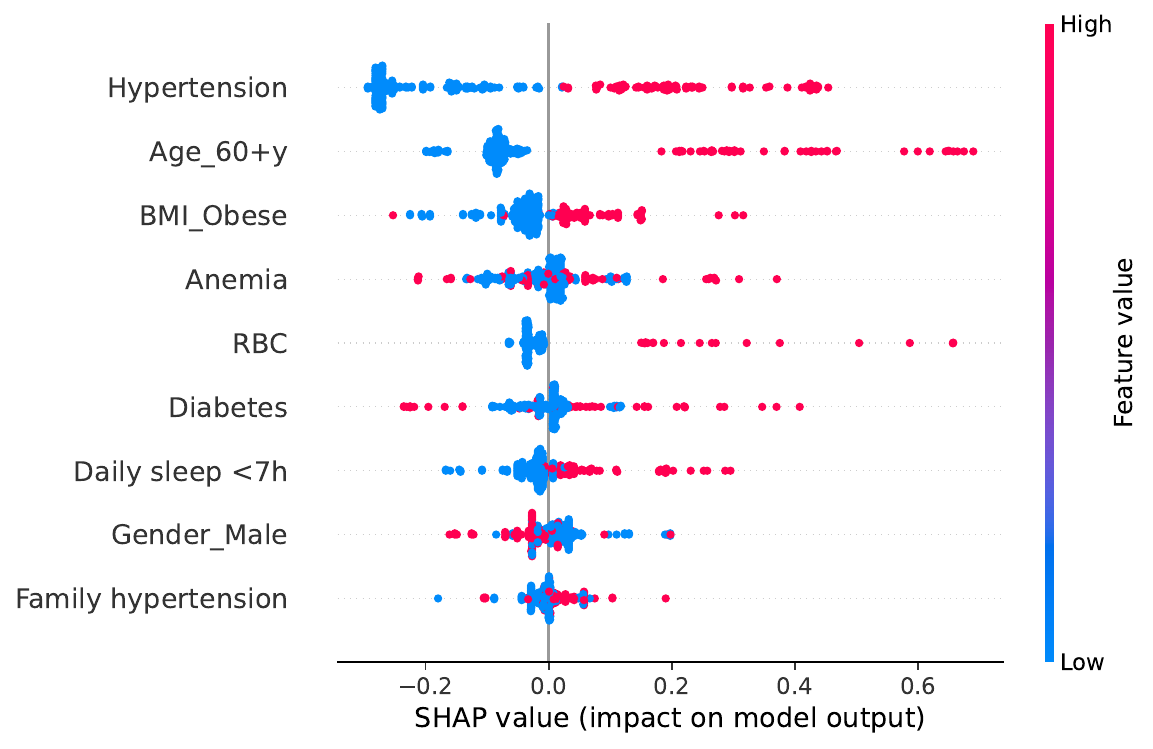}
\caption{SHAP summary plot illustrating the contribution of the best-performing S1 feature set to the Decision Tree model's predictions for CKD detection.}
    \label{fig:shap-summary}
\end{figure}

To enhance model transparency and clinical interpretability, we applied SHAP to analyze the contributions of the selected features in the best-performing model. Specifically, SHAP values were computed for the feature subset obtained using the RFECV+CB approach within the S1 setting, and interpreted using our best-performing DT classifier for CKD detection. Figure~\ref{fig:shap-summary} shows the resulting SHAP summary plot, where features are ranked along the Y-axis according to their mean absolute SHAP values, reflecting their overall importance and average impact on model predictions. The X-axis displays the SHAP values for individual instances, indicating both the magnitude and direction of each feature’s contribution. Positive SHAP values push the prediction toward CKD, whereas negative values push it toward non-CKD. Each point represents a single data instance, with colour denoting the magnitude of the one-hot encoded feature value (red indicating presence/``yes" and blue indicating absence/``no"). This explainability analysis provides insight into how key predictors influence decision-making within the model, supporting clinical trust and interpretability.

\begin{figure}[!ht]
    \centering
    \begin{subfigure}[b]{0.45\linewidth} 
        \centering
        \includegraphics[width=\textwidth]{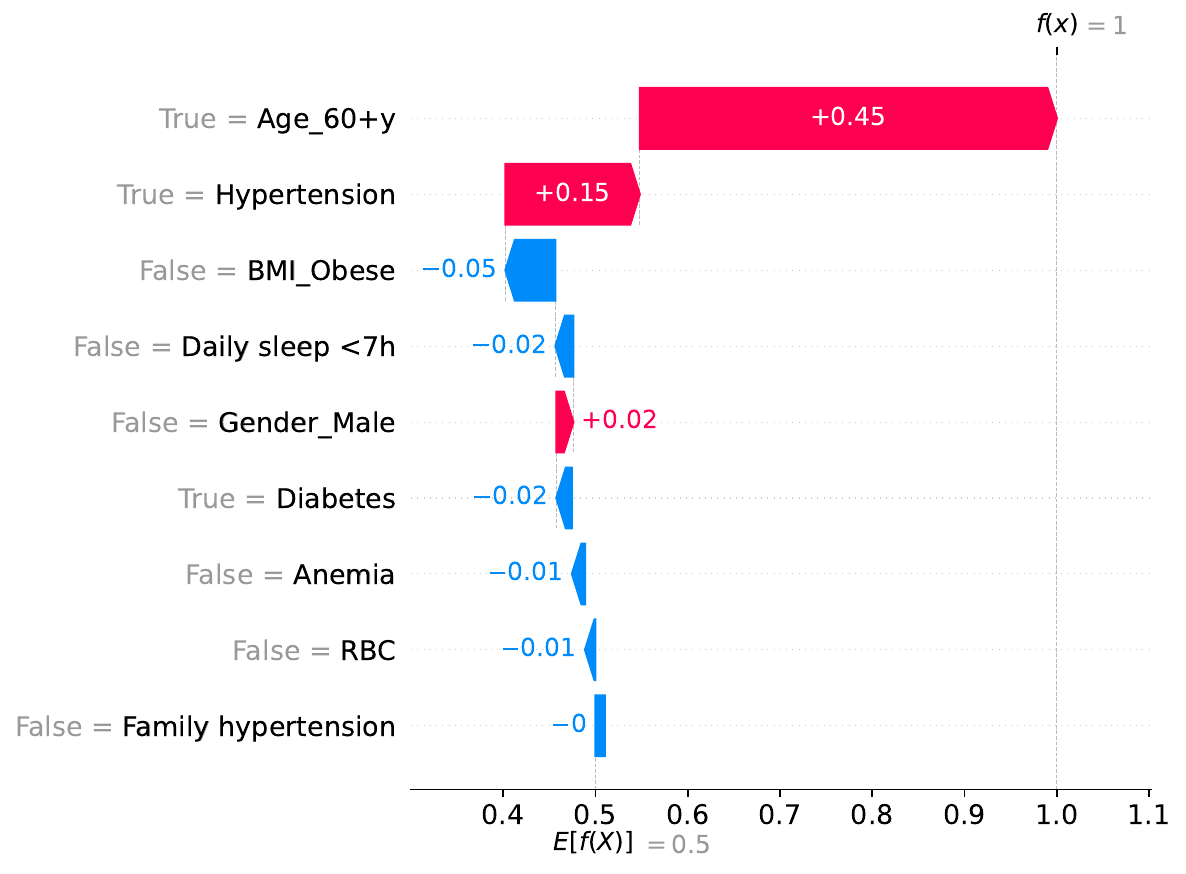}
        \caption{}
        \label{fig:shap-ckd}
    \end{subfigure}
    \begin{subfigure}[b]{0.54\linewidth}
        \centering
        \includegraphics[width=\textwidth]{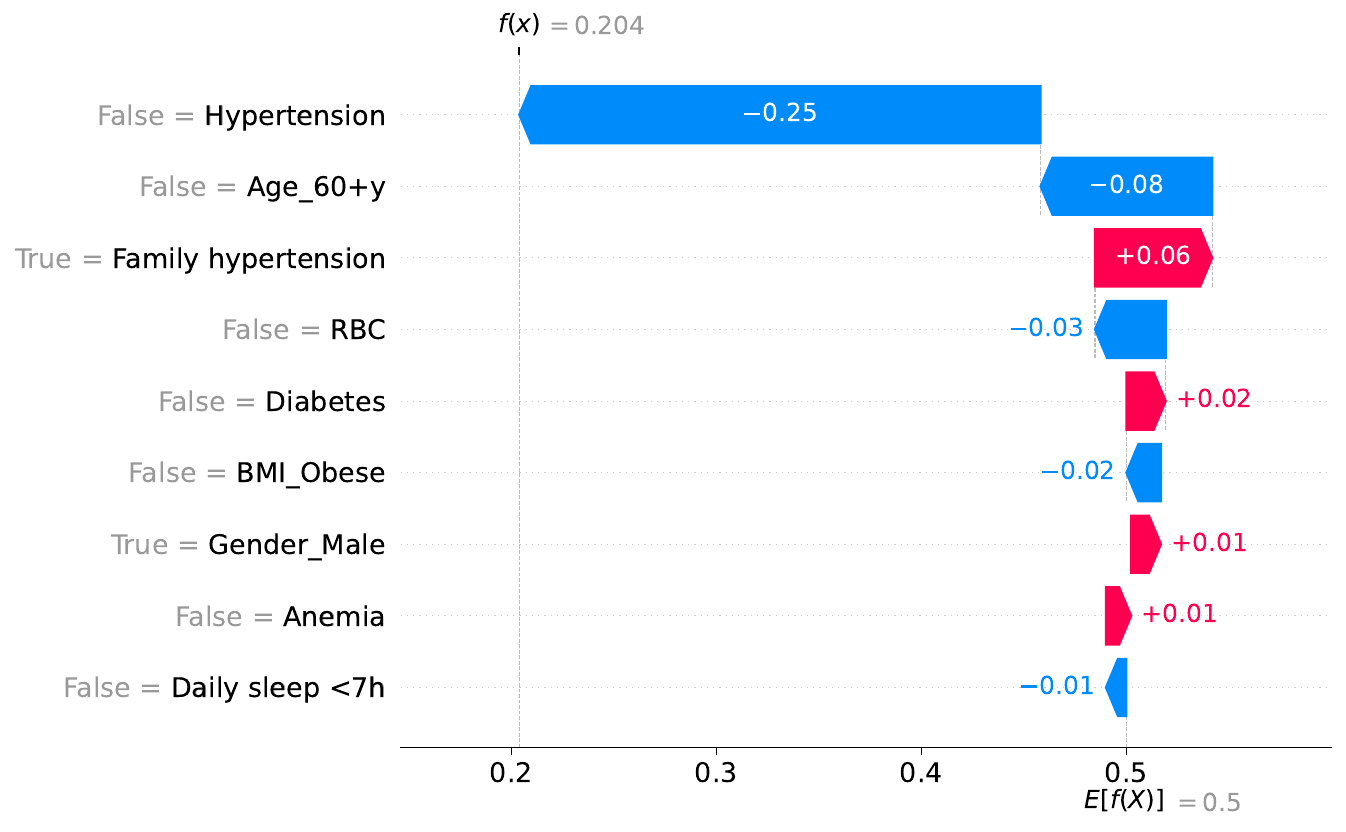}
        \caption{}
        \label{fig:shap-non-ckd}
    \end{subfigure}
    
    \caption{SHAP waterfall plot for a correctly classified (a) CKD and (b) Non-CKD case, illustrating the feature values contributing to the model prediction.}
    \label{fig:shap-sample}
\end{figure}

Figure~\ref{fig:shap-sample} presents SHAP waterfall plots for two representative prediction instances, a correctly classified CKD (Figure~\ref{fig:shap-ckd}) and non-CKD (Figure~\ref{fig:shap-non-ckd} case. Each plot illustrates how individual feature values contribute to shifting the model output from the baseline prediction (the expected value of the model) toward the final predicted probability. In Figure~\ref{fig:shap-ckd}, several high-risk indicators, including \texttt{Age\_\textit{60+y}}, \texttt{Hypertension}, and \texttt{Gender}, push the prediction strongly toward CKD, resulting in a final output close to 1. Features with negative contributions (e.g., absence of \texttt{BMI\_Obese}, \texttt{Anemia}, \texttt{RBC}) slightly offset the prediction but do not reverse the decision. In Figure~\ref{fig:shap-non-ckd}, the absence of key CKD risk factors (e.g., \texttt{Hypertension}, \texttt{Age\_\textit{60+y}}, \texttt{RBC}) drives the prediction toward the non-CKD class, producing a low predicted value. Some features, such as \texttt{Family hypertension} contribute small positive shifts, but the strong negative contributions dominate, keeping the final prediction below the decision threshold.

\section{Discussion}

This study provides several methodological and clinical insights that strengthen the evidence for using minimal, non-invasive features to develop robust and generalizable models for early CKD detection.

First, our findings show that the RFECV-based feature selection method consistently identifies smaller yet highly discriminative feature subsets compared with LASSO and the Mann–Whitney \textit{U} test. This observation is important because it demonstrates that wrapper-based approaches such as RFECV, particularly when combined with strong base learners such as CatBoost, are more effective in isolating the most informative predictors without inflating the feature space. The smaller feature subsets produced by RFECV preserve predictive strength while reducing redundancy, which is critical for practical deployment in community screening settings.



Second, while no single classifier consistently dominates across all feature sets, we observe that many models achieve closely comparable performance, reflecting the stable predictive structure in the data. However, AdaBoost emerges as the most frequently top-performing classifier (Figure~\ref{fig:model-count}). This suggests that boosting-based ensemble methods may be particularly well-suited for integrating heterogeneous, mixed-type clinical features when predicting CKD.

Third, although our focus was on early-stage CKD detection (stages 1–3), the model demonstrates excellent performance in detecting advanced-stage CKD as well. The TH dataset, which comprises only CKD stages 3–5, achieves an exceptional sensitivity of 98.21\% using the S2 feature subset. This robustness across disease stages suggests that the selected features capture fundamental CKD risk patterns regardless of severity. Moreover, successful external validation on datasets from India (UCI-2015) and the UAE (TH) indicates that our approach generalizes reasonably well across ethnically diverse populations in South Asia and the Middle East.

Fourth, the comparison with state-of-the-art CKD screening tools reveals that our models not only outperform existing methods in accuracy, sensitivity, and AUC-ROC but also rely on fewer and more readily available features. Tools such as SCORED, \citet{kshirsagar2008simple}, \citet{kshirsagar2008simple}, and \citet{kearns2013predicting} often require broader or more specialized input variables (see Table~\ref{tab:sota}), some of which may not be accessible in low-resource settings. In contrast, our optimized S1 and S2 subsets achieve superior performance with simpler, more scalable feature requirements, highlighting the practical advantage of our methodology for population-level CKD screening.

Finally, this study has not yet undergone clinical deployment. Prospective validation in real-world clinical or community settings will be a crucial next step to evaluate usability, acceptability, workflow integration, and impact on early CKD detection outcomes. Such studies will also help identify behavioural, operational, or socio-cultural factors that influence adoption.

In this study, individuals with advanced CKD (stage 4–5) were excluded from the primary analytic dataset because the primary objective was to develop models for community-based early-stage CKD screening (stages 1–3). In the TangailBD dataset, this exclusion involved a small number of participants (stage 4: n = 5; stage 5: n = 1). While this design focuses the model on early disease detection, it does not prevent the identification of advanced CKD cases in practice. Individuals with later-stage CKD typically exhibit stronger risk signals and would still be expected to be classified as high-risk. This is supported by the external validation results, where the model achieved high sensitivity on datasets containing more advanced CKD cases. In real-world deployment, the proposed framework is intended as a screening tool to identify individuals at elevated CKD risk, after which confirmatory clinical evaluation and laboratory testing would be used to determine disease stage and guide treatment.

A key limitation of this study is the relatively modest size and geographic scope of the primary dataset (TangailBD), which was collected from a specific region in Bangladesh. While the dataset is community-based and appropriate for the study objective, its limited sample size and regional focus may affect the robustness and representativeness of the learned models. Although external validation was conducted on three independent datasets to assess generalizability, further evaluation on larger and more diverse community-based populations is necessary to confirm the stability and broader applicability of the proposed framework.

In addition, although the proposed models demonstrate strong predictive performance, their effectiveness may be influenced by variations in data distribution and feature availability across populations, highlighting the importance of further validation under diverse real-world conditions.

\section{Conclusion}
This study presents a comprehensive and data-driven framework for early CKD detection using machine learning, demonstrating that strong predictive performance can be achieved with minimal, non-invasive, and readily obtainable features. Through rigorous feature selection, extensive model evaluation, comparison with state-of-the-art tools, and validation across three independent external datasets, our findings highlight the robustness, generalizability, and clinical relevance of the proposed approach. Notably, the S2 feature subset, requiring only two non-laboratory variables, achieves performance comparable to full feature sets, underscoring its potential for use in community-based and resource-limited settings where laboratory testing is not feasible. The explainability analysis further enhances transparency and trust by revealing clinically meaningful feature contributions. Overall, this work provides a strong foundation for future development of scalable CKD screening solutions. Further studies evaluating implementation feasibility, cost-effectiveness, and integration within community health systems will be required before real-world deployment can be considered.

\section*{Summary Table}
\noindent What was already known on the topic:
\begin{enumerate}
    \item Early-stage chronic kidney disease (CKD) is typically asymptomatic, and conventional diagnosis relies on pathology tests such as serum creatinine, eGFR, and ACR—tests that are often inaccessible or unaffordable in many low-resource settings.

    \item Several community-based CKD screening tools exist (e.g., SCORED), but most were developed in high-income or non–South Asian populations, or primarily target CKD stage $\ge3$ rather than early-stage disease.

    \item Existing machine learning studies on CKD have relied heavily on the UCI-2015 dataset and/or commonly include pathology-test-based features, including serum creatinine or eGFR, introducing information leakage or making them unsuitable for community-level screening in resource-limited environments.
\end{enumerate}

\noindent What this study adds to our knowledge:
\begin{enumerate}
    \item Developed the first community-based, early-stage CKD screening framework tailored to Bangladesh and the broader South Asian context, using only non-laboratory features that are feasible to collect in rural and peri-urban settings.

    \item Established a rigorous feature selection pipeline using ten complementary methods, leading to robust, generalizable, and easily obtainable features that can achieve strong predictive performance, outperforming larger feature sets and several existing clinical screening tools, including World Health Organization (WHO)-endorsed SCORED.



    \item Conducted extensive external validation across three independent datasets (from India, UAE, Bangladesh), showing the strong generalizability of the selected feature sets and demonstrating applicability across diverse ethnic and geographic populations.

    \item Provided transparent model interpretability using SHAP, revealing clinically meaningful feature contributions and increasing trustworthiness for potential deployment in community screening programmes.
\end{enumerate}

\section*{CRediT authorship contribution statement}
\textbf{Muhammad Ashad Kabir:} Conceptualization, Methodology, Software, Investigation, Validation, Formal analysis, Visualization, Project administration, Writing - Original Draft, Writing - Review \& Editing. \textbf{Sirajam Munira:} Investigation, Formal analysis, Validation, Writing - Original Draft. \textbf{Dewan Tasnia Azad:} Conceptualization, Investigation, Validation, Writing - Review \& Editing. \textbf{Saleh Mohammed Ikram:} Conceptualization, Investigation, Validation, Writing - Review \& Editing. \textbf{Mohammad Habibur Rahman Sarker:} Conceptualization, Data Curation, Validation, Writing - Review \& Editing. \textbf{Syed Manzoor Ahmed Hanifi:} Conceptualization, Project administration.

\section*{Declaration of competing interest}
The authors declare that they have no known competing interests that could have appeared to influence the work reported in this paper.

\section*{Data Availability Statement}
Of the four datasets used in this study, three are publicly available and can be accessed through their respective repositories as cited in the manuscript. The ICDDR,B dataset cannot be made publicly available due to ethical and institutional restrictions. Data requests for ICDDR,B can be directed to: \texttt{habibur.rahman@icddrb.org}.

\appendix
\setcounter{table}{0}
\section{Performance metrics definition}\label{app:performance:metric}
\paragraph{Sensitivity (Recall)}  
Sensitivity measures the proportion of actual CKD cases correctly identified by the classifier. It is a critical metric in medical diagnostics because missing positive cases can lead to delayed or missed treatment. Sensitivity is defined as:
\begin{equation}
\text{Sensitivity}_{\text{CKD}} = \frac{TP}{TP + FN},
\end{equation}
where $TP$ and $FN$ denote true positives and false negatives, respectively.

\paragraph{Balanced Accuracy}  
Accuracy alone is misleading for imbalanced datasets. Balanced accuracy mitigates this issue by averaging the recall of both positive and negative classes, ensuring performance assessment independent of class distribution. It is defined as:
\begin{equation}
\text{Balanced Accuracy} = \frac{1}{2} \left( \frac{TP}{TP + FN} + \frac{TN}{TN + FP} \right),
\end{equation}
where $TN$ and $FP$ denote true negatives and false positives, respectively.

\paragraph{F1 Score (CKD)}  
The F1 score emphasises the harmonic balance between precision and recall for the CKD class, making it suitable when false negatives and false positives are both undesirable. It is computed as:
\begin{equation}
\text{F1}_{\text{CKD}} = 2 \times \frac{\text{Precision}_{\text{CKD}} \times \text{Sensitivity}_{\text{CKD}}}{\text{Precision}_{\text{CKD}} + \text{Sensitivity}_{\text{CKD}}},
\end{equation}
where  
\begin{equation}
\text{Precision}_{\text{CKD}} = \frac{TP}{TP + FP}.
\end{equation}

\paragraph{Precision (macro)}  
Macro precision computes the unweighted mean precision across both classes, giving equal importance to CKD and non-CKD regardless of class size. This provides an unbiased measure of per-class correctness in the presence of imbalance:
\begin{equation}
\text{Precision (macro)} = \frac{1}{2} \left( \text{Precision}_{\text{CKD}} + \text{Precision}_{\text{Non-CKD}} \right).
\end{equation}
where  
\begin{equation}
\text{Precision}_{\text{Non-CKD}} = \frac{TN}{TN + FN}.
\end{equation}

\paragraph{AUC-ROC}  
The area under the receiver operating characteristic curve evaluates the classifier’s ability to discriminate between CKD and non-CKD cases across all possible decision thresholds. AUC-ROC is threshold-independent and robust to class imbalance, making it a widely recommended metric in clinical risk prediction. Formally,
\begin{equation}
\text{AUC-ROC} = \int_{0}^{1} TPR(FPR^{-1}(x)) \, dx,
\end{equation}
where $TPR$ is the true positive rate and $FPR$ is the false positive rate.


\section{CKD Screening Tools}\label{app:tools}
Table~\ref{tab:sota} provides a comparative overview of the feature sets used by established CKD screening tools alongside those used in this study (S1 and S2). For each tool, the table lists the country of origin, the CKD stages the tool is designed to detect, and the specific features utilised across five common domains: socio-demographic, lifestyle and habit, medical history, clinical examination, and pathology. This comparison highlights the substantial differences in feature coverage between models.
\begin{table}[!ht]
    \centering
    \caption{Features comparison with state-of-the-art (SOTA) CKD screening tools}
    \label{tab:sota}
    \resizebox{1\textwidth}{!}{
    \begin{tabular}{ll llcllc}
        \hline
        \multirow{2}{*}{Tool}  & \multirow{2}{*}{\makecell{Country}}  &\multirow{2}{*}{\makecell[c]{CKD\\[-4pt]stage}}
& \multicolumn{5}{c}{Features} \\ \cline{4-8}
         &  &
&  Socio-demo & \makecell[c]{Lifestyle\\[-4pt]\& habit} & Medical history & Clinical exam & Pathology \\
        \hline
        \hline
        SCORED~\citep{bang2007development} & USA   &3--5 
& Age, Gen & -- & HT, DM, PVD, CVD, HF & Ane, Ptn & -- \\
        \citet{kshirsagar2008simple} & USA   & 3--5 
& Age, Gen & -- & HT, DM, PVD, CVD, HF & Ane & -- \\
        \citet{thakkinstian2011simple} & Thailand   &1--5 
& Age & -- & DM, KS & Ane & -- \\
        \citet{kearns2013predicting} & England   &3--5 
& Age, Gen, Eth, Dep & Smk & HT, DM, PVD, CVD, HF, IHD, Str & BP & -- \\
        \citet{kwon2019simple}  & S. Korea  &3--5 & Age, Gen & -- & HT, DM, CVD & Ane, Ptn & -- \\
        \hline
This study (S1) & Bangladesh & 1--3 & Age, Gen & Daily sleep & HT, DM, Family HT & Ane, BMI & RBC\\
This study (S2) & Bangladesh & 1--3 & Age & Daily sleep & HT, DM &Ane & -- \\
        \hline

        \multicolumn{8}{l}{\footnotesize\makecell[l]{%
        Gen - Gender, Eth - Ethnicity, Dep - Deprivation Score, PVD - Peripheral Vascular Disease, CVD - Cardiovascular Disease, Ane - Anemia, BP - Blood Pressure, Ptn - Dipstick proteinuria,\\ 
        HF - Heart Failure, KS - Kidney Stone, IHD - Ischaemic Heart Disease, Str - Stroke, HT - Hypertension, DM - Diabetes Mellitus, Smk - Smoking, BMI - Body Mass Index,\\
        RBC - Presence of red blood cells in urine, Daily sleep - Daily sleep $<7$ hours
        }} \\

    \end{tabular}
    }
\end{table}

These tools predominantly rely on socio-demographic indicators and medical history variables, with limited or no use of lifestyle, clinical, or pathology features. In contrast, the proposed S1 and S2 models incorporate a more diverse and nuanced set of predictors, including daily sleep patterns, family history, and—in the case of S1—pathology indicators such as RBC. The table also shows that the models in this study target early CKD stages (1–3), whereas many established tools are designed for later stages (3–5).

\section{Extended Results}
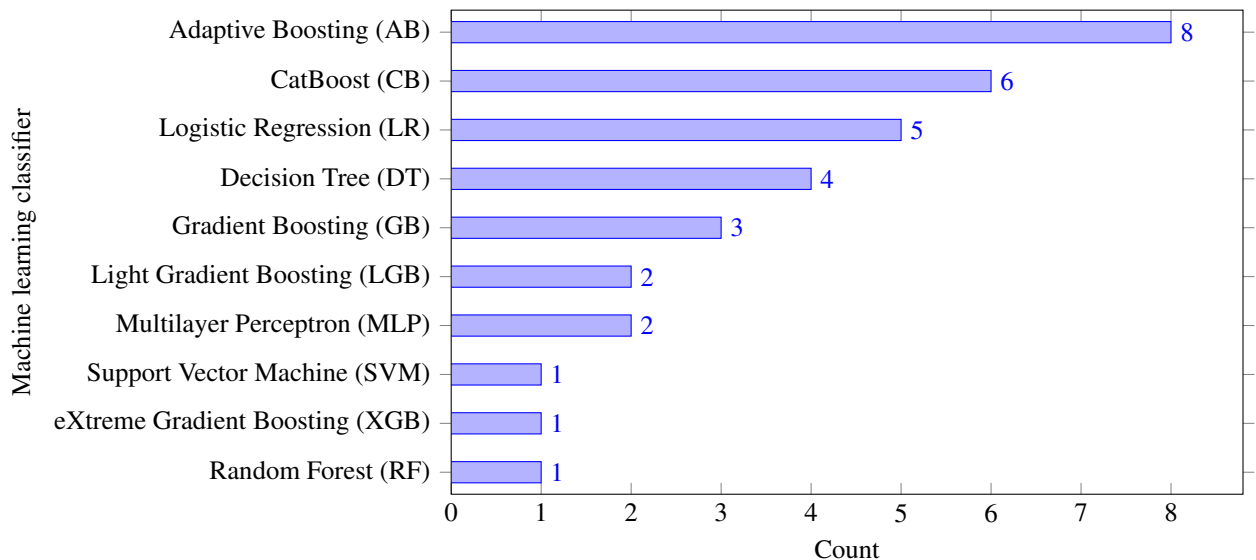
\begin{figure}[!ht]
\centering
\begin{tikzpicture}
\begin{axis}[
    xbar,
    xlabel={Count},
    ylabel={Machine learning classifier},
    symbolic y coords={AB,CB,LR,DT,GB,LGB,MLP,SVM,XGB,RF},
    yticklabels={
        Adaptive Boosting (AB),
        CatBoost (CB),
        Logistic Regression (LR),
        Decision Tree (DT),
        Gradient Boosting (GB),
        Light Gradient Boosting (LGB),
        Multilayer Perceptron (MLP),
        Support Vector Machine (SVM),
        eXtreme Gradient Boosting (XGB),
        Random Forest (RF)
    },
    ytick=data,
    width=12cm,
    height=8cm,
    xmin=0,
    bar width=8pt,
    nodes near coords,
    nodes near coords align={horizontal},
    enlarge y limits=0.05,
    y dir=reverse,   
]
\addplot coordinates {
    (8,AB)
    (6,CB)
    (5,LR)
    (4,DT)
    (3,GB)
    (2,LGB)
    (2,MLP)
    (1,SVM)
    (1,XGB)
    (1,RF)
};
\end{axis}
\end{tikzpicture}
\caption{Distribution of best-performing machine learning models identified across all experiments using different feature subsets. The counts reflect how often each classifier achieved the highest balanced accuracy within a given feature configuration.}
\label{fig:model-count}

\end{figure}

\pagebreak

\bibliographystyle{elsarticle-num-names} 
\bibliography{reference}

\end{document}